\documentclass[10pt]{article} %

\usepackage[accepted]{tmlr}
\def\month{12}  %
\def\year{2026} %
\def\openreview{\url{https://openreview.net/forum?id=xNXsvdAMbJ}}

\usepackage[utf8]{inputenc} %
\usepackage[T1]{fontenc}    %
\usepackage{hyperref}       %
\usepackage{url}            %
\usepackage{booktabs}       %
\usepackage{amsmath}        %
\usepackage{amssymb}        %
\usepackage{amsfonts}       %
\usepackage{nicefrac}       %
\usepackage{microtype}      %
\usepackage{xcolor}         %
\usepackage{placeins}       %
\usepackage{graphicx}       %
\usepackage{wrapfig}        %
\usepackage{subcaption}     %
\usepackage{pifont}         %
\newcommand{\cmark}{\ding{51}}

\graphicspath{{figures/}}

\newif\ifshowrev
\showrevfalse
\newcommand{\rev}[1]{\ifshowrev\textcolor{blue}{#1}\else#1\fi}

\title{Bigger Text Encoders Can Hurt CLIP Zero-Shot Performance}

\author{\name Samir Char\thanks{Corresponding author. \textbf{Code}: \url{https://github.com/samirchar/clip-asymmetry}.} \email samirchar@microsoft.com \\
      \addr Microsoft
      \AND
      \name Carles Domingo-Enrich \email carlesd@microsoft.com \\
      \addr Microsoft Research
      \AND
      \name Randall Balestriero \email randall\_balestriero@brown.edu \\
      \addr Brown University}

\begin{document}

\maketitle

\begin{abstract}

Contrastive Language-Image Pretraining (CLIP) is a building block of many machine learning applications.
Scaling laws have guided resource allocation for large-scale training, yet prior work treats total CLIP model size as a single variable,
without exploring how the capacity split between encoders impacts downstream performance.
Here, we train multiple CLIP models with different vision and text encoder sizes \rev{on CC3M and CC12M}, 
revealing that for most vision encoders, there is an optimal text encoder size beyond which zero-shot performance degrades---%
even as total parameter count increases. 
Exploiting this behavior yields configurations that match the zero-shot performance of the standard ViT-B/16 architecture with up to 55\% fewer parameters \rev{and 48\% fewer FLOPs}. 
We further show that this degradation stems from overfitting induced by the oversized text encoder \rev{(i.e., better training fit but worse zero-shot transfer)}, 
and that using modality-specific weight decay coefficients not only recovers but improves performance across all degraded configurations. 
A geometric analysis reveals a trade-off in which scaling the text encoder improves embedding uniformity but worsens cross-modal alignment; we
further show that these metrics are predictive of zero-shot performance \rev{across our tested architectures}.
We hope these findings motivate CLIP architectures and training methods that counteract this degradation\rev{, a prerequisite for scaling CLIP more reliably across vision and text encoder capacity splits}.

\end{abstract}

\section{Introduction} 

Contrastive Language-Image Pretraining (CLIP)~\citep{radford2021clip} trains paired vision and text encoders with a
contrastive objective on large-scale image--text data, yielding representations with remarkable zero-shot transfer.
Since its introduction, CLIP has become a standard building block for a wide range of applications, including image generation~\citep{saharia2022imagen,ramesh2022dalle2,esser2024stablediff3},
generative vision-language models~\citep{liu2023llava,awadalla2023openflamingo}, video understanding~\citep{xu2021videoclip,wang2024videoclipxl}, 
and image segmentation~\citep{kirillov2023sam,ravi2025sam2}. As these downstream systems inherit CLIP encoders or use their embeddings,
the question of how to design and scale CLIP architectures has broad practical consequences.

\begin{figure}[t]
  \centering
  \includegraphics[width=\linewidth]{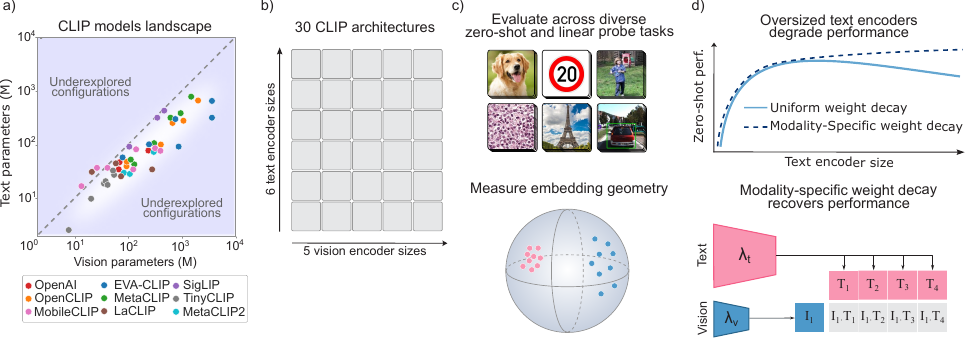}
  \caption{Naively scaling CLIP can degrade downstream performance.
(a)~Popular CLIP models by vision (x-axis) and text encoder (y-axis) size ($n{=}45$; points jittered).
Most models cluster just below the diagonal, leaving most vision-text capacity splits unexplored.
(b)~A grid of 30 architectures spanning five vision and six text encoder sizes,
each pre-trained on CC3M and CC12M.
(c)~Every configuration is evaluated across 30 zero-shot and 31 linear-probe tasks, and its embedding geometry is quantified
via alignment, uniformity, modality gap, and RankMe.
(d)~\rev{Schematic.} Counterintuitively, under standard training (solid) zero-shot performance
rises, peaks and then degrades as the text encoder grows with the vision encoder fixed;
modality-specific weight decay (dashed) removes the degradation by applying separate
coefficients $\lambda_v$ and $\lambda_t$ to the encoders.}
  \label{fig:overview}
\end{figure}

A common strategy for improving CLIP performance is to scale up the model, yet the allocation of capacity between the vision 
and text encoders remains an implicit, unexamined design choice. Most work training CLIP-style models at multiple scales presents
a menu of architectures without specifying a scaling recipe or justifying the capacity split 
between encoders (Fig.~\ref{fig:overview}(a))~\citep{radford2021clip,zhai2023siglip,alabdulmohsin2024siglip2,cherti2023openclip,gadre2023datacomp,xu2023metaclip,chuang2025metaclip2}. 
Existing efforts to study CLIP scaling laws treat total model size as a single variable~\citep{cherti2023openclip,li2023clipa,li2023clipa2},
without explicitly testing whether one encoder should be scaled more aggressively than the other. 
Yet the two modalities are inherently asymmetric: text is a compact, abstract signal whereas images are high-dimensional 
and redundant, and prior multimodal learning literature shows that modalities can overfit and generalize at different rates~\citep{wang2021multimodal}.
Scaling both encoders equally may not merely be sub-optimal but actively harmful.
To our knowledge, no prior work independently varies each encoder's capacity to isolate its marginal 
contribution to downstream performance, its impact on overfitting, or its effect on the geometry of the learned embedding space.

In this work, we conduct a controlled study of how independently scaling CLIP's vision and text encoders affects zero-shot and linear probe performance.
We train a full grid of 30 models, crossing five vision encoders (from 10M to 352M parameters) with six text encoders (from 3M to 145M parameters) on CC3M and CC12M, and evaluate each across 40 downstream datasets (Fig.~\ref{fig:overview}(b,c)).
Our experiments yield three main contributions:
 \begin{enumerate}
   \item We find that for zero-shot tasks CLIP's vision and text encoders should not be scaled equally. For most vision encoders there is an optimal
     text encoder size beyond which performance degrades (solid curve, Fig.~\ref{fig:overview}(d)). This pattern is absent in the linear
     probe setting, where larger text encoders consistently improve or, at worst, saturate performance.
     Exploiting this asymmetry, we identify two architectures that match the
     zero-shot performance of the standard ViT-B/16 architecture while
     using 30--55\% fewer total parameters.
   \item We show that this degradation stems from a capacity mismatch between the encoders that manifests as asymmetric 
    overfitting from an oversized text encoder. Using modality-specific weight decay to \rev{regularize each encoder separately}, we find that, for a given vision encoder,
    increasing the text encoder's weight decay recovers---and even improves---zero-shot performance,
     \rev{matching or} surpassing the best architecture with the same vision encoder (dashed curve, Fig.~\ref{fig:overview}(d)). 
     Raising only the vision encoder's weight decay instead degrades performance,
     while the conventional practice of decaying both encoders equally underperforms
     decaying the text encoder alone in most scenarios, or performs comparably.
   \item A geometric analysis of the representations through the
     lens of alignment and uniformity~\citep{wang2020alignment} shows that
     scaling the text encoder improves the uniformity of image and text embeddings
     while degrading image-text alignment. Raising the text encoder's weight decay mitigates this effect, recovering alignment while preserving
     or even improving image uniformity.

 \end{enumerate}

\section{Related Work} 

\paragraph{Evidence for asymmetric encoder scaling.}
 Prior work provides indirect evidence that the capacity split
 between CLIP's encoders matters. \citet{fan2023laclip} train CLIP at \rev{three}
model sizes by scaling only the vision encoder
while keeping the text encoder constant, reporting improved performance
over the original scaling recipe. EVA-CLIP-18B~\citep{sun2024evaclip18b} shows remarkable performance pairing a 17.5B-parameter vision
 encoder with a 695M-parameter text encoder.
 \citet{zhai2022lit} show that freezing the vision encoder entirely and
 training only the text side can yield strong zero-shot performance. Closest to our work, SmolVLM~\citep{allal2025smolvlm} runs a 2D grid
 search over vision and language backbone sizes for small VLMs, finding that the capacity allocation between the vision encoder and text decoder is critical for compute-efficient performance.
 However, SmolVLM studies decoder-based vision-language models rather than
 contrastive pre-training. 
 None of these works systematically isolates the effect of each encoder's size in a controlled contrastive pre-training setting. Our work addresses this gap.

 \paragraph{CLIP scaling laws.}
Inspired by language model scaling laws~\citep{hoffmann2022chinchilla},
 \citet{cherti2023openclip} demonstrated that CLIP's zero-shot performance follows a power-law
 relationship with model size, dataset size, and samples seen. Similarly,
 CLIPA~\citep{li2023clipa} and CLIPAv2~\citep{li2023clipa2} identify an inverse scaling law for CLIP
 training: using larger vision and text encoders enables training with
 aggressive token masking ratios, reducing compute while maintaining
 competitive performance. However, these studies treat the vision and text
 encoders as a single pool of parameters and do not decompose scaling effects
 between them. Our work complements these efforts by studying how scaling each encoder independently
 affects downstream performance, but we do not derive specific scaling laws.

\paragraph{Heterogeneous weight decay in deep learning.}
Different parts of a network may overfit at different rates, 
and should therefore be regularized differently. To address this, prior work adjusts weight
decay over time, per layer, per module, and even per
parameter~\citep{ghiasi2023awd,ishii2017layer,he2025alphadecay,nakamura2019adadecay}. 
Notable examples are AdaDecay~\citep{nakamura2019adadecay}, which scales the decay per parameter from gradient
norms, and AlphaDecay~\citep{he2025alphadecay}, which sets module-level decay from the
heavy-tailedness of the weight spectral density.
For CLIP and its variants, 
researchers have explored asymmetric per-encoder weight decay when one encoder 
is initialized from pretrained weights~\citep{zhai2023siglip,zhai2022lit,sun2023evaclip}.
Such an encoder is either kept frozen or trained with a smaller or zero weight decay to mitigate catastrophic forgetting.
Our work differs in that we vary weight decay across modalities while training both encoders from scratch (i.e., random initialization),
 using it as a deliberate \rev{regularizer} to mitigate modality-specific overfitting and improve zero-shot performance.

\paragraph{Embedding geometry in contrastive learning.}
The quality of contrastive representations has been characterized through alignment and
 uniformity~\citep{wang2020alignment}: alignment measures the closeness of
 embeddings from positive pairs, while uniformity measures how well
 representations are spread across the unit hypersphere. Both properties are necessary
 for good contrastive representations, yet there is an inherent tension between them.
 Separately, RankMe~\citep{garrido2023rankme} provides a label-free estimate of the
 effective dimensionality of embeddings that
 correlates with downstream transfer performance. We use alignment, uniformity, and RankMe as diagnostic tools to explain the scaling behaviors observed in
 our experiments; to our knowledge, no prior work has studied how
 independently scaling the vision and text encoders affects the geometry of
 the embedding space.

\section{Methodology}

\subsection{Asymmetric architecture grid}
To study how encoder size affects performance, we size our encoders to span both below and above
the standard ViT-B/16 architecture~\citep{radford2021clip,cherti2023openclip,fan2023laclip,xu2023metaclip},
which uses vision and text encoders of 86M and 63M parameters, respectively.
We define five vision encoders from 10M to 352M parameters,
six text encoders from 3M to 145M parameters,
and create a 2D grid of all possible combinations (Fig.~\ref{fig:overview}(b); Table~\ref{tab:architectures}).
This yields 30 configurations totaling 13M--497M parameters.

To scale the encoders, we follow prior work~\citep{radford2021clip,cherti2023openclip,gadre2023datacomp,xu2023metaclip,chuang2025metaclip2}, 
which scales the number of layers and attention heads, and derives the width and MLP hidden dimension from these choices.
We keep the patch size (16), input resolution ($224 \times 224$), context length (77), vocabulary size (49,408),
and shared embedding dimension (512) constant across all models to isolate the effect of scaling the encoders.
All architectures are implemented and trained using the OpenCLIP library~\citep{cherti2023openclip} and the original CLIP loss~\citep{radford2021clip}.
Throughout the paper, we denote each configuration by its vision--text encoder pair
(e.g., Base--Tiny uses a Base vision encoder with a Tiny text encoder).

All 30 architectures are pre-trained on two datasets of different scale: Conceptual Captions 3M (CC3M)~\citep{sharma2018cc3m}
and Conceptual Captions 12M (CC12M)~\citep{changpinyo2021cc12m}, resulting in a total of 60 models. We follow the same training hyperparameters
as in~\citet{fan2023laclip}, which are detailed in Appendix~\ref{sec:impl_details}.

\begin{table}[t]
  \caption{Vision and text encoder architectures in the model grid: five vision encoders spanning 10M--352M parameters crossed with six text encoders spanning 3M--145M, for 30 configurations.
  All models share a vocabulary size of 49,408, context length of 77 tokens, patch size of 16, input resolution of $224 \times 224$, and a shared embedding dimension of 512.}
  \label{tab:architectures}
  \centering
  \small
  \begin{minipage}[t]{0.48\linewidth}
    \centering
    (a) Vision Encoders\\[2pt]
    \begin{tabular}{lrrrr}
      \toprule
      Name & Layers & Width & Heads & Params \\
      \midrule
      Atto & 12 & 256 & 4 & 10M \\
      Tiny & 10 & 640 & 10 & 50M \\
      Base & 12 & 768 & 12 & 86M \\
      Giant & 20 & 896 & 14 & 194M \\
      Colossal & 22 & 1152 & 18 & 352M \\
      \bottomrule
    \end{tabular}
  \end{minipage}
  \hfill
  \begin{minipage}[t]{0.48\linewidth}
    \centering
    (b) Text Encoders\\[2pt]
    \begin{tabular}{lrrrr}
      \toprule
      Name & Layers & Width & Heads & Params \\
      \midrule
      Femto & 1 & 64 & 1 & 3M \\
      Atto & 3 & 128 & 2 & 7M \\
      Nano & 6 & 256 & 4 & 18M \\
      Tiny & 10 & 384 & 6 & 37M \\
      Base & 12 & 512 & 8 & 63M \\
      Giant & 15 & 768 & 12 & 145M \\
      \bottomrule
    \end{tabular}
  \end{minipage}
\end{table}

\subsection{Evaluation protocols}\label{sec:eval_protocols}

For a comprehensive assessment of downstream performance, 
we evaluate across 40 downstream datasets---28 for zero-shot classification and retrieval,
and 31 for linear probe classification. Most of these datasets are part
of the DataComp~\citep{gadre2023datacomp} benchmark, and we follow the same protocol and evaluation metrics.
The datasets include ImageNet~\citep{russakovsky2015imagenet} along with five ImageNet distribution shifts~\citep{wang2019imagenetsketch,recht2019imagenetv2,hendrycks2021natural,hendrycks2021manyfaces},
the 19 VTAB tasks~\citep{zhai2019vtab},
Flickr30k~\citep{young2014flickr30k} and MSCOCO~\citep{lin2014mscoco} for retrieval tasks, 
and other common classification tasks 
used in the original CLIP paper~\citep{radford2021clip}. Because Flickr30k and MSCOCO retrieval are each evaluated in both the image-to-text and text-to-image directions, the 28 zero-shot datasets correspond to 30 zero-shot tasks. Most datasets
are used for both zero-shot and linear probe evaluation, but some are only 
used for one of the two protocols as detailed
in Appendix~\ref{sec:downstream_datasets}.
We report the average performance across all datasets using
the preferred metric for each task (Table~\ref{tab:eval-datasets}), rescaled to a percentage $\in[0,100]$. 
Absolute differences in average performance are reported in percentage points (pp).
We also examine how the scaling patterns manifest across selected dataset groups (Table~\ref{tab:dataset-groups}).

For zero-shot classification, we adopt the same prompt templates used by~\citet{radford2021clip}.
Hand-crafted templates convert each class label into multiple text prompts; the
average embedding across all prompts for a class yields the class representation.
Predictions are made by computing the cosine similarity between the image embedding
and each class representation. 

In the linear probe setting, we train a linear classifier on top
 of the frozen image features extracted from the vision encoder, 
 without using any text prompts. It is trained for 10 epochs using 
 Adam~\citep{kingma2015adam} optimizer with neither weight decay nor learning
 rate decay, following~\citet{caron2021dino,he2022mae}.

We leverage the CLIP Benchmark library~\citep{cherti2023clipbenchmark} 
to implement both zero-shot and linear probe evaluation protocols.

To assess significance across tasks, we report the number of tasks on which 
one configuration outperforms another out of the non-tied total (n/N), along
with the $p$-value of a two-sided sign test.
\rev{Because the sign test assumes independent tasks, we also report a grouped version that first averages the differences within two groups of possibly correlated tasks: the seven ImageNet-derived datasets, and the four image-to-text and text-to-image retrieval tasks on MSCOCO and Flickr30k (21 units in total; Tables~\ref{tab:wd-sign-tests} and~\ref{tab:degraded-fixed}).}
Significance across random seeds is assessed with a paired $t$-test on the per-seed differences,
with the direction (one- or two-sided) indicated where reported.
\rev{The configurations compared in these tests were selected after inspecting the full grid.}

\section{Larger text encoders hurt zero-shot performance}
\label{sec:zs_scaling}

\begin{figure}[t]
  \centering
  \includegraphics[scale=1]{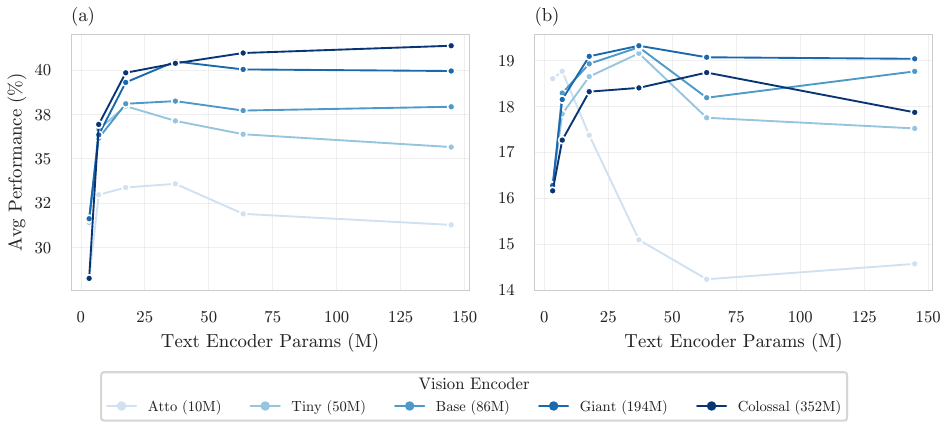}
  \caption{Beyond an optimal size, larger text encoders hurt zero-shot performance.
  Average zero-shot performance (y-axis; higher is better), averaged over 30 zero-shot tasks, as a function of text encoder size (x-axis), colored by vision encoder, for models pre-trained on (a)~CC12M and (b)~CC3M.}
  \label{fig:zs_scaling}
\end{figure}

To understand the impact of independently scaling the vision and text encoders, we train all 30 configurations on both CC3M and CC12M,
and evaluate their zero-shot performance across the 30 zero-shot tasks (Fig.~\ref{fig:zs_scaling}). 
Four patterns emerge consistently across both pre-training datasets.
First, for most vision encoders there is an optimal text encoder size beyond which zero-shot performance degrades.
Even the smallest degradation, observed between Base--Tiny and Base--Giant, is statistically significant across
three random seeds (one-sided paired $t$-test $p{=}0.04$, Cohen's $d_z{=}1.901$; Table~\ref{tab:seed-variance}).
Second, once the text encoder is sufficiently large, increasing the vision encoder becomes the dominant driver of performance,
and its benefit amplifies when paired with larger text encoders.
Third, the magnitude or onset of degradation varies with the vision encoder size, with the smallest
vision encoders showing more pronounced degradation than larger ones\rev{, and the degradation being most severe when the text encoder is as large as or larger than the vision encoder}.
Fourth, the optimal text encoder size tends to increase with the vision encoder,
but at a slower rate than the vision encoder itself. 
On CC12M the largest vision encoder shows no degradation, though even larger text encoders might eventually induce it.
In all our experiments, the smallest text encoder (Femto; 3M parameters) leads to poor and unpredictable 
performance for both datasets and evaluation settings
(Fig.~\ref{fig:zs_scaling}).
These patterns are consistent across 
all examined groups (Fig.~\ref{fig:zs-by-group}), with
the degradation being most pronounced for fine-grained tasks (Fig.~\ref{fig:zs-by-group}(c)).

One exception to these trends is attributable to overfitting on the smaller CC3M dataset.
The largest vision encoder (Colossal) does not achieve the best zero-shot performance on CC3M, 
consistent with known overfitting effects on smaller datasets~\citep{cherti2023openclip,fan2023laclip}
and with the learning curves (Figs.~\ref{fig:learning_curves_colossal} and~\ref{fig:learning_curves_base}).

In contrast to zero-shot, scaling the text encoder does not degrade linear
probe performance (Figs.~\ref{fig:lp_scaling} and~\ref{fig:lp-by-group}) and only exhibits monotonic improvement or saturation at worst.

\begin{figure}[t]
  \centering
  \includegraphics[scale=1]{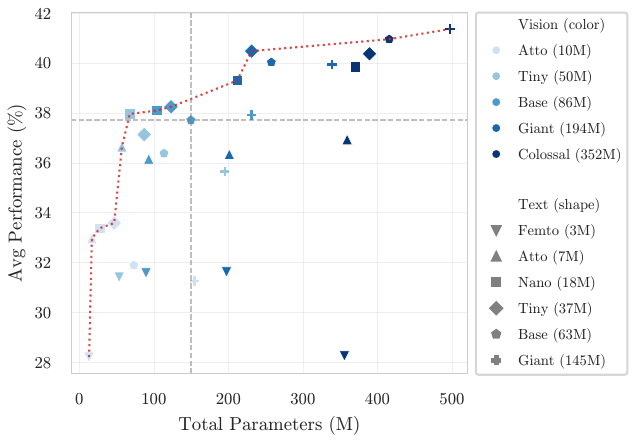}
  \caption{Exploiting the vision--text asymmetry yields configurations matching ViT-B/16 zero-shot performance with 30--55\% fewer parameters.
  Average zero-shot performance (y-axis; higher is better) versus total model parameters (x-axis) on CC12M for all 30 architectures.
  Color encodes the vision encoder and marker shape encodes the text encoder; the red dotted line traces the Pareto frontier and the gray dashed lines mark the Base--Base (ViT-B/16) baseline.}
  \label{fig:pareto} 
\end{figure}

Given that bigger text encoders are not always better, we ask whether it is possible to match the zero-shot performance 
of the base ViT-B/16 CLIP architecture (Base--Base) with fewer total parameters. The Pareto frontier (Fig.~\ref{fig:pareto}) 
is dominated by configurations with smaller text encoders, and models with similar zero-shot performance can have markedly different total parameters.
Concretely, we find that while Base--Base reaches 37.7\% average zero-shot performance, 
the Base--Nano and Tiny--Nano models \rev{match} it (38.1\% and 37.9\%) with \rev{30.7\%} and 54.7\% fewer parameters, respectively.
\rev{Besides the reduction in parameter count, compared to the baseline Base--Base architecture, Base--Nano needs 12.6\% fewer FLOPs (98\% of its FLOPs are in the vision encoder), 17\% less peak memory, 49\% lower text-encoder latency, 4.2$\times$ higher text-encoder throughput, and a 16\% shorter training step. Tiny--Nano is even more efficient because of its smaller vision encoder (48\% fewer FLOPs, 41\% less peak memory and a 42\% shorter training step; Appendix~\ref{sec:efficiency}).}

\section{Parameter allocation between encoders matters beyond total count}
\label{sec:allocation}

\begin{figure}[t]
  \centering
  \includegraphics[width=\linewidth]{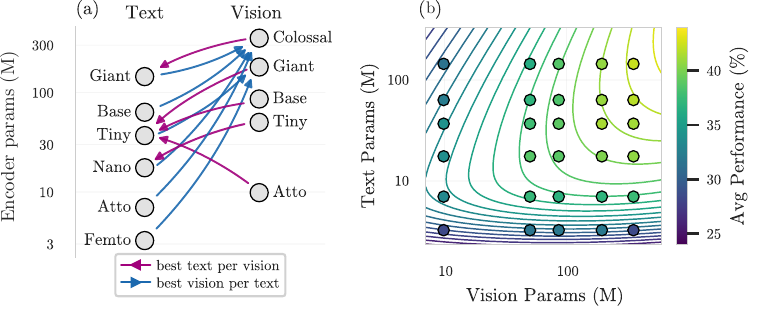}
  \caption{The vision--text parameter split matters beyond the total parameter count.
  (a)~Each node represents an encoder positioned on the y-axis according to its size.
  Left-to-right arrows (blue) point from each text encoder to the best vision encoder paired with it in terms of average zero-shot performance; right-to-left arrows (magenta) point from each vision encoder to the best text encoder.
  (b)~Predicted average zero-shot performance (higher is better) from the fitted saturating-ceiling regression with a capacity-mismatch penalty (Eq.~\ref{eq:ceiling}), as a function of the vision (x-axis) and text (y-axis) encoder size.
  Contour lines trace the regression's iso-performance levels and the 30 architectures are overlaid as points colored by their observed average zero-shot performance.}
  \label{fig:split_matters}
\end{figure}

Analyzing the CC12M training runs reveals that CLIP architectures with the best zero-shot performance have markedly different 
vision and text encoder capacities.
Across the grid the zero-shot-optimal text encoder for every vision encoder is less than 40M parameters for all vision encoders except Colossal (352M), 
  whose best text encoder has 145M parameters (Fig.~\ref{fig:split_matters}(a)).
In contrast, the optimal vision encoder for any text encoder has at least 194M parameters, 
  and for four of the six text encoders the best vision encoder has 352M parameters.
  For the largest text encoders the optimal vision encoder size is right-censored, since zero-shot
performance is still rising at our largest 352M vision encoder. Even larger vision encoders might yield better results.
We observe that in multiple cases the optimal size for one encoder increases as the other grows.

To quantify the relationship between model capacity and zero-shot performance, 
we fit multiple parametric models on the CC12M results using the vision, text, and total parameter counts as independent variables (Table~\ref{tab:scaling-model-comparison}).
Although total parameter count positively correlates with zero-shot performance (Spearman $r{=}0.62$, $p{<}0.001$), 
a linear-log regression on total parameters explains only a small fraction of the variance (Adjusted-$R^2{=}0.312$; leave-one-out cross-validated, or LOOCV, $R^2{=}0.240$).
Fitting a linear-log model on the vision parameters, text parameters and an interaction term yields a substantially better fit (Adjusted-$R^2{=}0.693$; LOOCV $R^2{=}0.543$), 
confirming that how capacity is allocated between encoders matters beyond the total budget.

Remarkably, the best fit comes from a simple nonlinear model that treats the two encoders asymmetrically and penalizes a capacity mismatch 
(LOOCV $R^2{=}\rev{0.946}$).
Writing $N_v$ and $N_t$ for the vision and text encoder parameter counts (in millions), we fit the form
\begin{equation}
  \hat{y}(N_v, N_t) = \underbrace{\left(\beta_0 + \beta_1\,\log_{10} N_v\right)}_{\text{vision-set ceiling}}\,
  \underbrace{\left(1 - e^{-k\,N_t}\right)}_{\text{text saturation}}\,
  \underbrace{\exp\!\left(-c\left(\log_{10}\tfrac{N_v}{N_t} \rev{{}- r_0}\right)^2\right)}_{\text{capacity-mismatch penalty}},
  \label{eq:ceiling}
\end{equation}
where $\hat{y}$ is the predicted average zero-shot performance. The term $\beta_0 + \beta_1\log_{10} N_v$ is a linear-log ceiling that is scaled by a
saturating factor $1 - e^{-k N_t}\in[0,1]$ in the text size and further scaled by the mismatch penalty $\exp(-c\,(\log_{10}(N_v/N_t) \rev{{}- r_0})^2)\in(0,1]$\rev{, which equals 1 (no penalty) when the vision encoder is $10^{r_0}$ times the size of the text encoder}.
Removing the mismatch penalty fits substantially worse (LOOCV $R^2{=}0.826$), underscoring the importance of appropriate encoder capacity.
Visualizing this fit over the encoder-size plane shows that, once the text encoder is adequately sized, zero-shot performance benefits more from scaling vision,
and that the degradation from oversized text encoders is strongest for the smallest vision encoders and gradually vanishes as the vision encoder grows (Fig.~\ref{fig:split_matters}(b)).
\rev{Additionally, the fitted $r_0{=}0.33$ means that the penalty is smallest when the vision encoder is about $2.1\times$ the size of the text encoder (95\% CI $[1.3\times, 3.4\times]$).}
We present these findings as an interpretable description of our data rather than a definitive scaling law. 
Characterizing the full functional form over a wider range of text encoder sizes is a promising direction to predict the optimal parameter split 
given a total parameter budget, akin to~\citet{hoffmann2022chinchilla} for language models.

\section{Modality-specific weight decay mitigates text-encoder overfitting}
\label{sec:mswd}

\begin{figure}[t]
  \centering
  \includegraphics[width=\linewidth]{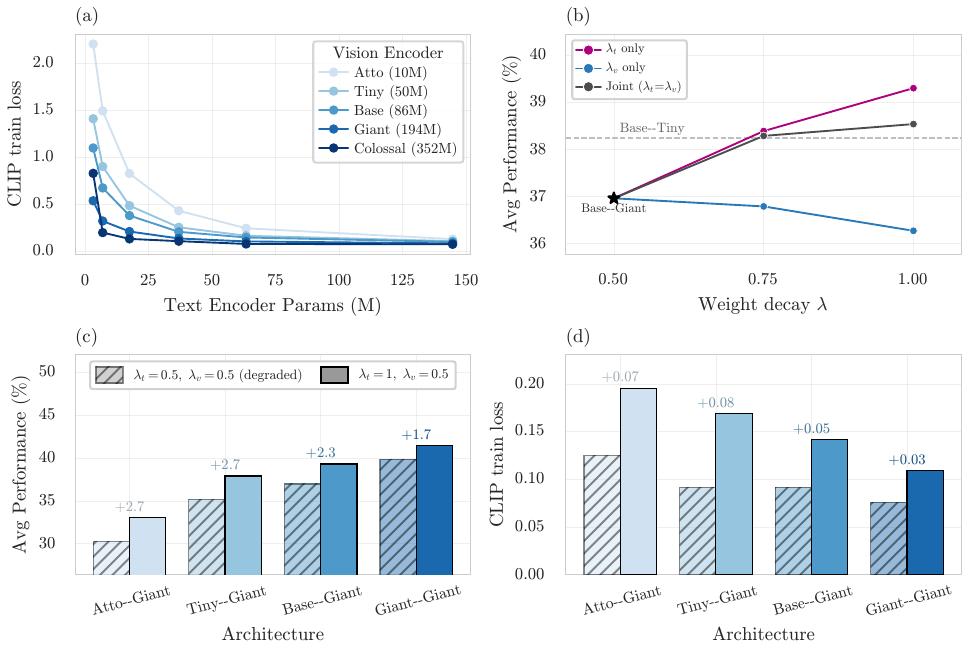}
  \caption{CLIP demands modality-specific encoder capacity.
  (a)~Final-epoch CLIP training loss on CC12M (y-axis) versus text-encoder size (x-axis), colored by vision encoder.
  (b)~Average zero-shot performance (y-axis; higher is better) for the Base--Giant configuration as weight decay is increased on the text encoder only ($\lambda_t$), the vision encoder only ($\lambda_v$), or both jointly, all emanating from the Base--Giant baseline with joint weight decay of 0.5.
  The dashed line marks the best configuration using the Base vision encoder (Base--Tiny).
  (c)~Average zero-shot performance (y-axis; higher is better) for each vision encoder paired with the oversized Giant text encoder (x-axis), comparing the degraded configuration ($\lambda_t{=}0.5$, $\lambda_v{=}0.5$; hatched bars) with the fixed configuration ($\lambda_t{=}1$, $\lambda_v{=}0.5$; solid bars).
  (d)~Final-epoch CLIP training loss on CC12M (y-axis) for the same configurations as in (c), comparing the degraded configuration (hatched) with the fixed configuration (solid).}
  \label{fig:mswd}
\end{figure}

Prior work has shown that different modalities overfit and generalize at different rates, so
training them jointly with a single optimization strategy is typically sub-optimal~\citep{wang2021multimodal}. 
\rev{Information-bottleneck views of CLIP also suggest that unnecessary modality-specific information is discarded only implicitly, through limited capacity and the regularization inherent in optimization~\citep{ji2025cibr}, so oversized encoders may retain irrelevant noise unless properly regularized (Appendix~\ref{sec:ib_discussion}).}
Motivated by \rev{these observations}, we hypothesized that in CLIP the text encoder is more prone to overfitting---especially when training with web-scale noisy captions---than the vision encoder, 
and that this asymmetry drives the zero-shot degradation observed in Section~\ref{sec:zs_scaling}.

To assess whether optimal text encoders are small because larger ones overfit, we compute CLIP's contrastive
loss on the CC12M training set.
Growing the text encoder monotonically lowers the training loss for every vision encoder (Fig.~\ref{fig:mswd}(a)),
even as zero-shot performance degrades once the text encoder exceeds its optimum, as shown previously in Figure~\ref{fig:zs_scaling}(a).
These results show that, for a fixed vision encoder, growing the text encoder beyond its optimum leads to overfitting\rev{, by which we mean that zero-shot transfer declines while the fit to the training data keeps improving~\citep[cf.][]{radford2021clip}}.
\rev{However, when we retrain the models with a portion of CC12M held out, the degraded configurations actually achieve better retrieval recall on this held-out dataset than the configurations with the best average zero-shot performance across our evaluation suite (Atto, Tiny and Base vision encoders; Fig.~\ref{fig:heldout_recall}(a--f)), suggesting that oversized text encoders may specifically hurt out-of-distribution generalization.}

Driven by these findings, we tested whether modality-specific weight decay improves zero-shot performance by independently 
\rev{regularizing each encoder}.
 This strategy splits the model parameters into a vision and a text group, assigning each
  separate weight-decay coefficients ($\lambda_v$ and $\lambda_t$), held constant throughout training (see Appendix~\ref{sec:impl_details}). 
Our baseline is Base--Giant, 
whose oversized text encoder yields degraded zero-shot performance under the default $\lambda_v{=}\lambda_t{=}0.5$.
We then train three families of variants, raising the weight decay to 0.75 and 1.0 on 
(i) the text encoder only ($\lambda_t$), (ii) the vision encoder only ($\lambda_v$), or (iii) both encoders jointly, as the conventional practice (Table~\ref{tab:wd-ablation-full}). 
Higher weight decay values ($\lambda{=}1.5$) caused learning collapse in all cases and are therefore excluded (Fig.~\ref{fig:wd_collapse}).
Figure~\ref{fig:mswd}(b) shows that raising the text
weight decay consistently improves average zero-shot performance by 1.4 pp at $\lambda{=}0.75$ and 2.3 pp at $\lambda{=}1.0$, 
whereas only raising the vision weight decay degrades it
by up to 0.7 pp at 1.0 (text-only wins 25/30 tasks at $\lambda{=}1$, $p{<}0.001$).
Raising both jointly balances these effects for a net gain,
but a smaller one than text-only decay (38.5\% vs.\ 39.3\% at $\lambda{=}1$; text-only wins 23/30 tasks, $p{=}0.005$). 
Notably, text-only weight decay (39.3\%) surpasses the Base--Tiny configuration (38.2\%),
the best model using the Base vision encoder.
\rev{The gain from the same weight-decay intervention decreases, and may even lead to degradation, as the text encoder size decreases (Table~\ref{tab:wd-controls}).}
This asymmetry holds in the Giant--Giant configuration where the vision encoder is the larger of the two backbones (194M vs.\ 145M parameters; \rev{Fig.~}\ref{fig:wd_ablation_giant}): 
raising the text encoder's weight decay improves performance while raising the vision encoder's degrades 
it (text-only wins on 23/30 tasks, $p{=}0.005$).

Next we retrained and evaluated the zero-shot performance of the four degraded architectures paired 
with the Giant text encoder (Atto, Tiny, Base, and Giant vision encoders) using modality-specific weight decay with $\lambda_t{=}1.0$ and $\lambda_v{=}0.5$.
This intervention consistently improves zero-shot performance for every configuration, from +2.7 pp for Atto--Giant 
to +1.7 pp for Giant--Giant, with the gain shrinking as the vision encoder grows and there is less overfitting to mitigate (all with sign-test $p{<}0.05$; Fig.~\ref{fig:mswd}(c); Table~\ref{tab:degraded-fixed}).
Strikingly, the performance of the Giant--Giant configuration rises to $41.5\%$, 
\rev{exceeding} the best Giant-vision architecture by 1.0 pp (Giant--Tiny; sign test $p{=}0.016$) 
and matching the strongest model in the entire grid, Colossal--Giant ($41.4\%$; sign test $p{=}0.585$), 
despite a vision encoder roughly half its size (194M vs.\ 352M parameters).
While modality-specific weight decay improves zero-shot performance, it increases the training loss in all degraded configurations (Fig.~\ref{fig:mswd}(d)).
\rev{Using dropout to regularize only the text encoder has a similar effect at the baseline weight decay, improving all four degraded configurations by 2.1 to 3.3 pp (Appendix~\ref{sec:dropout}).}

Together, these results point to three principles: CLIP benefits from a larger vision than text encoder;
both must be sized appropriately to avoid text-encoder overfitting or vision-encoder underfitting;
and the capacity mismatch between the two encoders modulates this effect.

\section{Encoder scaling and regularization reshape embedding geometry}
\label{sec:geometry}

\begin{figure}[t]
  \centering
  \includegraphics[scale=1]{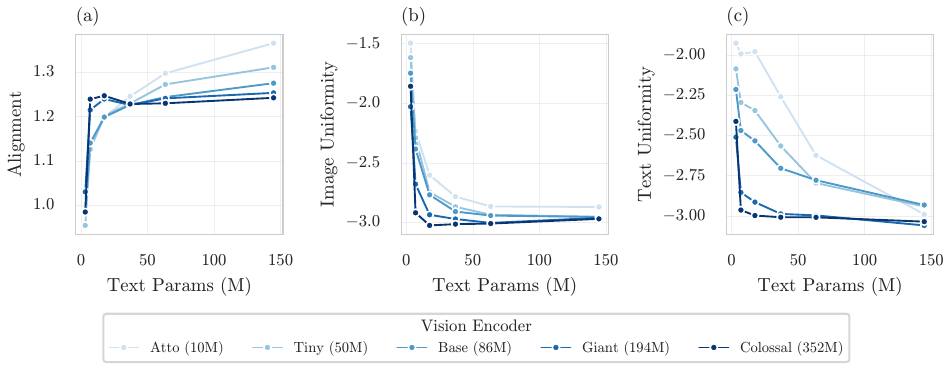}
  \caption{Scaling the text encoder improves embedding uniformity but worsens image--text alignment.
  (a)~Alignment, (b)~image uniformity, and (c)~text uniformity (lower is better for all three; y-axes) of MSCOCO embeddings ($n{=}5{,}000$ image--caption pairs) from the CC12M pre-trained models, plotted as a function of text encoder size (x-axis) and colored by vision encoder.}
  \label{fig:geometry_text}
\end{figure}

\begin{wrapfigure}{r}{0.45\textwidth}
  \centering
  \includegraphics[scale=1]{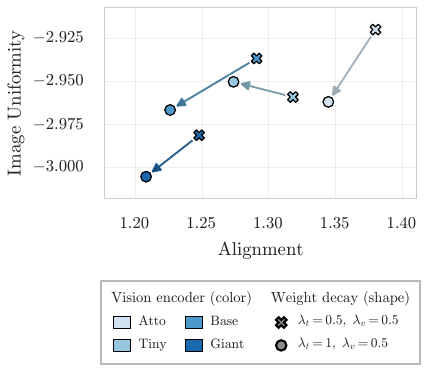}
  \caption{Modality-specific weight decay improves image-text alignment.
  Embedding alignment (x-axis) versus image uniformity (y-axis) on MSCOCO (lower is better for both), with arrows pointing from the degraded (crosses) to the fixed (circles) configuration using modality-specific weight decay. Colors denote the vision encoder, all paired with the Giant text encoder.}
  \label{fig:degraded_fixed_geometry}
\end{wrapfigure}
To understand how asymmetric scaling affects the embedding space, we compute the alignment and uniformity~\citep{wang2020alignment} of the
image and text embeddings. Alignment measures the expected distance between matched (positive) image--text pairs, and uniformity measures how evenly the
embeddings spread over the unit hypersphere. Both are framed as losses, so lower is better. We focus on CC12M models, where the scaling patterns
are not confounded by the overfitting seen on CC3M, and compute these metrics on the 5,000 images
in the MSCOCO~\citep{lin2014mscoco} Karpathy split~\citep{karpathy2015deep}, using only the first caption for each image. 

Larger text or vision encoders tend to improve both image and text embedding uniformity (Figs.~\ref{fig:geometry_text}(b,c) and \ref{fig:geometry_vision}(b)).
Scaling the text encoder improves image uniformity at a faster rate than the vision encoder. 
Additionally, as the text encoder grows, text uniformity keeps rising while image uniformity saturates at the Base text encoder (63M parameters).
Even though scaling the text encoder improves image uniformity, it worsens image-text alignment, 
but the degree of this degradation decreases with larger vision encoders (Fig.~\ref{fig:geometry_text}(a)).
Increasing the vision encoder worsens alignment when paired with small text encoders, but improves it when paired with the \rev{largest} (Fig.~\ref{fig:geometry_vision}(a)).

Beyond scaling, modality-specific weight decay also reshapes the embedding geometry. 
Applying it for the degraded configurations improves image--text alignment
while maintaining or improving image uniformity, on both MSCOCO (Fig.~\ref{fig:degraded_fixed_geometry})
and Flickr30k (Fig.~\ref{fig:degraded_fixed_geometry_grid}(c)), though typically at the expense of text uniformity (Fig.~\ref{fig:degraded_fixed_geometry_grid}(b,d)).

To evaluate whether uniformity and alignment are predictive of the average zero-shot performance,
we separately extract features from the MSCOCO Karpathy split and Flickr30k, and fit multiple linear regressions on combinations of the three metrics (alignment, image uniformity, and text uniformity).
We find that a model trained on alignment and image uniformity has the most predictive power (MSCOCO LOOCV $R^2{=}0.810$; Flickr30k LOOCV $R^2{=}0.877$; Table~\ref{tab:geometry-prediction}). 
These results are compatible with prior work showing
that these metrics correlate with downstream performance~\citep{wang2020alignment,alabdulmohsin2024siglip2}. 
Using the fitted regression model we plot the predicted zero-shot performance landscape as a function of alignment and image uniformity (Fig.~\ref{fig:geometry_landscape}),
showing that the best models cluster together, while degraded models worsen their alignment without significant improvement in image uniformity. 
To test generalization, we fit the regression in a leave-one-vision-out fashion, where we train on all but one vision encoder,
and then predict the average zero-shot performance of the left-out vision encoder (Fig.~\ref{fig:geometry_lovo}). 
The regression captures the overall trend, including the peak and degradation, 
although it also predicts a degradation for the largest vision encoder (Colossal) that is not observed in practice.
\rev{Since uniformity and alignment are measured on MSCOCO and Flickr30k, which are also included in the zero-shot evaluation suite, we repeat the same analysis excluding these two datasets from the evaluation and observe that the results remain practically unchanged (MSCOCO LOOCV $R^2{=}0.805$; Flickr30k LOOCV $R^2{=}0.875$; Table~\ref{tab:geometry-prediction-noleak}).}
These results highlight a practical implication: the alignment and uniformity measured on a small held-out dataset can estimate the expected zero-shot performance of a CLIP model
across a wide range of tasks \rev{within the range of architectures we study. Whether this predictive relationship extends to other pre-training datasets, larger models, or
different families of downstream tasks remains to be tested}.

\paragraph{Interpretation.} 
Because zero-shot inference needs both well-separated embeddings (uniformity) 
and aligned image--text spaces (alignment), scaling the text encoder creates 
a trade-off: it improves uniformity but often at the expense of alignment.
The saturation and degradation pattern we observe in zero-shot
performance is reflected in the embedding geometry, coinciding with the point at which alignment
degradation begins to outweigh uniformity gains---especially when the
vision encoder is not large enough.
This interpretation is consistent with
our largest vision encoder, the only one that does not degrade 
and for which using the largest text encoders does not meaningfully deteriorate alignment.
\rev{We treat alignment and uniformity as diagnostic correlates of the degradation, not as a causal account of it.}

In contrast, linear probe inference depends only on the vision encoder, 
where performance hinges on the linear separability of image embeddings rather than cross-modal alignment.
This aligns with our observation that larger text encoders improve image uniformity, and therefore the linear
separability of image embeddings.
Further analysis and details are provided in Appendix~\ref{sec:add_geometry}.

\section{Conclusion}

In this work, we presented a controlled study of asymmetric encoder scaling in CLIP, revealing that
 increasing the total number of parameters can consistently degrade zero-shot performance if capacity 
 is not split appropriately between the vision and text encoders. 
 Once the text encoder is large enough, performance is 
 driven by the vision encoder, and further increasing the text encoder degrades it.
This asymmetry lets us identify configurations that match the zero-shot performance of the standard ViT-B/16 
architecture with 30--55\% fewer total parameters. We show that the degradation stems from overfitting induced by the oversized text encoder \rev{(i.e., better training fit but worse zero-shot transfer)},
which manifests in the embedding space as a worsening of image-text alignment and improved uniformity.
Modality-specific weight decay recovers and even improves both performance and image-text alignment across all degraded architectures,
while preserving or improving uniformity. Crucially, conventional uniform weight decay helps less than modality-specific weight decay in most scenarios, or performs comparably, 
reinforcing the need for \rev{modality-specific regularization}.

There are several avenues for future work. Our findings motivate the development 
of additional training methods that mitigate the zero-shot degradation. We envision three main directions, including modality-specific regularization 
or optimization strategies, architectural modifications, and dataset improvements such as using longer and richer captions.
 Eliminating the degradation would enable more resilient scaling behavior where zero-shot
 performance increases monotonically with compute while being less sensitive to the encoder capacity split. 
 While it persists, identifying
 the optimal parameter allocation between encoders across a wider range of dataset scales and qualities
 could enable principled, parameter-efficient CLIP scaling recipes.
\rev{Likewise, characterizing the values of $\lambda_t$ and $\lambda_v$ that maximize downstream performance while maintaining training stability would facilitate the practical use of modality-specific weight decay.}
\rev{Another direction is downstream adaptation. We evaluate using zero-shot classification and linear probes, but CLIP models are often adapted with parameter-efficient methods such as prompt tuning~\citep{zhou2022coop}. Smaller models would make prompt tuning cheaper (Table~\ref{tab:efficiency}), but the resulting downstream performance is still an open question that could be tested. In addition, our evaluation only uses the final representations. Since some adaptation methods rely on the different information captured by different layers~\citep{khattak2023maple,xiao2026cascaded}, it is worth exploring how the quality and content of the representations at different depths change when selecting a smaller text encoder or applying modality-specific weight decay.}
Finally, generalizing these findings to multiple and heterogeneous modalities is an important next step. Researchers
could explore how each modality's data statistics inform how to build compute-optimal, large-scale multimodal models beyond vision and text.

\section{Limitations}

Our study has several limitations.
First, we train exclusively on CC3M and CC12M, but the degradation could diminish or disappear with larger pre-training datasets.
Second, we evaluate only on classification and retrieval tasks; the impact of asymmetric scaling may differ for other downstream applications such as generation and segmentation.
Third, we pre-train all models from scratch, whereas most practitioners initialize from pre-trained encoders, which may alter the dynamics we report.
Fourth, even though modality-specific weight decay proved effective in our experiments, in some cases it led to training instabilities and learning collapse.
Fifth, we cannot perform significance tests across multiple seeds for all analyses due to the high computational cost involved.
\rev{Sixth, all models are trained from scratch on datasets with predominantly English captions and are evaluated mainly on
classification and retrieval tasks. Because CC3M and CC12M differ in scale and curation simultaneously, our setup cannot
isolate the effect of caption quality from that of dataset size. Taken together with the limitations above, our results should be read as evidence that an
optimal capacity split exists, while the specific location of that optimum is tied to our setting and may shift with
pre-training scale, caption distribution, quality and language, training objective and recipe, and the composition of the evaluation suite.} \rev{We expect the optimum to slowly move toward larger text encoders as the data scales up and captions become cleaner and richer.}

\subsubsection*{Broader Impact Statement}
We do not anticipate significant risks of harm beyond those already associated with CLIP-style models.
\rev{Our models are trained on CC3M and CC12M, web-scraped alt-text datasets with known social biases, and likely inherit them; we do not assess whether encoder size or modality-specific regularization affects these biases.}
Our work shows that asymmetrically scaling CLIP's encoders can \rev{match} strong baselines with a 
fraction of the parameters.
This lowers the barrier to training CLIP from scratch under limited compute\rev{, as fewer total parameters can substantially reduce memory requirements}.
\rev{Configurations with smaller text encoders may also slightly reduce the environmental footprint, mainly during training and at inference when text embeddings cannot be cached.}

\makeatletter
\if@accepted
\subsubsection*{Author Contributions}
Conceptualization: S.C., C.D.E., R.B.;
Data Curation: S.C.;
Formal Analysis: S.C.;
Investigation: S.C.;
Methodology: S.C., C.D.E., R.B.;
Resources: C.D.E.;
Software: S.C.;
Supervision: C.D.E., R.B.;
Validation: S.C.;
Visualization: S.C.;
Writing -- Original Draft: S.C.;
Writing -- Review \& Editing: S.C., C.D.E., R.B.

\subsubsection*{Acknowledgments}
The authors thank Hannes Schulz, Sean Whitzell and Kevin K. Yang for assistance with Microsoft's compute resources; and Eric Zimmermann for helpful discussions.

\subsubsection*{Declaration of Interests}
The authors declare no competing interests.
\fi
\makeatother

\bibliographystyle{tmlr}
\bibliography{references}

\appendix
\renewcommand{\thefigure}{A\arabic{figure}}
\renewcommand{\thetable}{A\arabic{table}}
\setcounter{figure}{0}
\setcounter{table}{0}

\section{Implementation details}
\label{sec:impl_details}

\paragraph{Encoders}
Our baseline architecture corresponds to the ViT-B/16 from~\citet{cherti2023openclip}, pairing an 86M-parameter vision encoder with a 63M-parameter text encoder.
To construct encoder variants of different sizes, we follow the standard Transformer scaling rules used by prior work~\citep{dosovitskiy2021vit,radford2021clip,cherti2023openclip}.
We only scale the number of attention heads $h$ and the number of layers while keeping the per-head dimension fixed at 64, which determines the model width as $d_{\text{model}} = 64 \times h$.
The MLP hidden dimension is set to $4 \times d_{\text{model}}$.
The same rules apply to the vision and text encoders. The resulting configurations are listed in Table~\ref{tab:architectures}.

\paragraph{Hyperparameters}
We follow the same training recipes used by~\citet{fan2023laclip} for both the CC3M and CC12M datasets. Table~\ref{tab:hparams} provides an overview of the pre-training hyperparameters used for all models on each dataset.
We apply RandomResizedCrop with a scale range of $(0.5, 1.0)$ as the image augmentation.
All pre-training runs were conducted on four to eight machines equipped with 8 A100 GPUs each.

\begin{table}[htbp]
  \centering
  \caption{Pre-training hyperparameters for (a)~CC3M and (b)~CC12M datasets. All runs use the AdamW optimizer~\citep{loshchilov2019adamw}.}
  \label{tab:hparams}
  \begin{subtable}[t]{0.48\textwidth}
    \centering
    \caption{CC3M.}
    \label{tab:hparams_cc3m}
    \begin{tabular}{@{}ll@{}}
      \toprule
      Config & Value \\
      \midrule
      Batch size            & 8,192 \\
      Optimizer             & AdamW \\
      Learning rate         & $1\times10^{-3}$ \\
      Weight decay          & 0.5 \\
      Adam $\beta$          & $\beta_1,\beta_2=0.9,\,0.98$ \\
      Adam $\epsilon$       & $1\times10^{-8}$ \\
      Total epochs          & 40 \\
      Warmup epochs         & 1 (355 steps) \\
      LR schedule           & Cosine decay \\
      \bottomrule
    \end{tabular}
  \end{subtable}
  \hfill
  \begin{subtable}[t]{0.48\textwidth}
    \centering
    \caption{CC12M.}
    \label{tab:hparams_cc12m}
    \begin{tabular}{@{}ll@{}}
      \toprule
      Config & Value \\
      \midrule
      Batch size            & 8,192 \\
      Optimizer             & AdamW\\ 
      Learning rate         & $1\times10^{-3}$ \\
      Weight decay          & 0.5 \\
      Adam $\beta$          & $\beta_1,\beta_2=0.9,\,0.98$ \\
      Adam $\epsilon$       & $1\times10^{-8}$ \\
      Total epochs          & 35 \\
      Warmup epochs         & 1 (1,339 steps) \\
      LR schedule           & Cosine decay \\
      \bottomrule
    \end{tabular}
  \end{subtable}
\end{table}

\paragraph{\rev{Fixed hyperparameters across configurations}}
\rev{To isolate the effect of encoder capacity, we hold most settings fixed across configurations, including the projection dimension, the text context length, the batch size, the learning rate, the number of epochs and the default weight decay.
A fixed 512-dimensional shared embedding limits what either encoder can contribute to the contrastive objective, and a constant context length of 77 over a fixed vocabulary bounds the information available per caption.
Holding these fixed should at most saturate performance; we do not expect it to cause the degradation we observe.}

\rev{Fixing the batch size and the number of epochs fixes the number of updates and ensures that every sample is seen the same number of times.
Changing the batch size would also change the objective itself, since the contrastive loss is defined over the batch: larger batches provide more negatives.
Holding these settings constant is therefore necessary for a fair comparison.}

\rev{The optimal learning rate may change with encoder size, but this alone is unlikely to explain the degradation.
If the learning rate or other fixed settings were responsible, they should handicap both encoders; instead, scaling the vision encoder improves average zero-shot performance while scaling the text encoder degrades it (Fig.~\ref{fig:zs_scaling}). Still, comparisons under a fixed recipe cannot rule out such architecture-dependent optimization effects.}

\rev{Lastly, we follow the training recipe of LaCLIP~\citep{fan2023laclip}, which trains ViT-S/16, ViT-B/16 and ViT-L/16 on CC12M with the same hyperparameters, including the learning rate.
More broadly, using one learning rate across CLIP model sizes is standard at this scale: OpenCLIP's scaling study uses the same rate for ViT-B/32, B/16 and L/14 at a given batch size~\citep[Table~18]{cherti2023openclip}.}

\paragraph{Modality-specific weight decay}
Weight decay is applied per parameter group under AdamW. All one-dimensional
parameters are excluded from weight decay: 
bias from fully connected layers, gains and biases from layer normalization, the visual class embedding, and the logit scale.
The remaining parameters are split into two groups:
\begin{itemize}
  \item Vision: all remaining parameters used by the vision encoder, including the patch-embedding convolutions,
  positional embeddings, attention and MLP matrices, and the final projection matrix.
  \item Text: all remaining parameters used by the text encoder, including the token embeddings, 
  positional embeddings, attention and MLP matrices, and the final projection matrix.
\end{itemize}

Vision parameters are then optimized with a weight decay coefficient $\lambda_v$ and text parameters with a weight decay coefficient $\lambda_t$, which
are held constant throughout training. The baseline and joint-decay runs use $\lambda_v=\lambda_t$.

\section{Dataset details}

\subsection{Pre-training datasets}

We pre-train on two image-text datasets at different scales: CC3M~\citep{sharma2018cc3m} and CC12M~\citep{changpinyo2021cc12m}.
To mitigate the impact of link rot, we use archived versions of both datasets that were downloaded and stored on Hugging Face in 2021.\footnote{\url{https://huggingface.co/datasets/conceptual_captions} and \url{https://huggingface.co/datasets/conceptual_12m}}
These versions contain fewer samples than the original releases, which may lead to minor differences in performance relative to models trained on the complete datasets.

CC3M~\citep{sharma2018cc3m} contains 3.3 million image-text pairs collected from approximately 5 billion web pages, where captions are sourced from HTML alt-text attributes.
Our copy retains 2.9 million unique samples.

CC12M~\citep{changpinyo2021cc12m} follows a similar collection pipeline but applies less restrictive filtering, yielding 12.4 million image-text pairs that cover a broader range of visual concepts and topics.
The version we use contains 10.9 million samples.

For both datasets, all images were resized so that their shorter side equals 256 pixels.

\subsection{Downstream datasets}
\label{sec:downstream_datasets}

Table~\ref{tab:eval-datasets} lists the datasets used for zero-shot and linear probe evaluation, 
along with their task types, sizes, number of classes, main evaluation metrics,
and whether they are used for zero-shot (ZS) and/or linear probe (LP) evaluation.

For zero-shot evaluation, we exclude \rev{11} of the VTAB datasets because they are known to
perform no better than random guessing~\citep{gadre2023datacomp}. These include 
all \rev{nine} structured tasks \rev{(VTAB's eight, with dSprites location split into $x$ and $y$ position)}, the diabetic retinopathy task\rev{, and PatchCamelyon}. We also
exclude Face Expression Recognition (FER2013)~\citep{goodfellow2013fer2013} for the same reason.

In the linear probe protocol, we exclude Flickr30k~\citep{young2014flickr30k} and MSCOCO~\citep{lin2014mscoco} because they are 
retrieval datasets that are not compatible with linear probe evaluation.
The five ImageNet distribution 
shifts (ImageNet Sketch~\citep{wang2019imagenetsketch}, ImageNet v2~\citep{recht2019imagenetv2}, 
ImageNet-A~\citep{hendrycks2021natural}, ImageNet-O~\citep{hendrycks2021natural}, and ImageNet-R~\citep{hendrycks2021manyfaces}),  
Scene Understanding (SUN-397)~\citep{xiao2010sun}, and ObjectNet~\citep{barbu2019objectnet} are
also excluded due to incompatibility with the CLIP Benchmark
library~\citep{cherti2023clipbenchmark}.

\begin{table}[htbp]
  \caption{Evaluation tasks in the benchmark suite. Columns ZS and LP indicate whether each dataset is used for zero-shot and linear probe evaluation, respectively; a checkmark (\cmark) marks inclusion.}
  \label{tab:eval-datasets}
  \centering
  \resizebox{\textwidth}{!}{%
  \begin{tabular}{llllrrl cc}
    \toprule
    Task type & Dataset & Task & Test size & Classes & Main metric & ZS & LP \\
    \midrule
    Classification & Caltech-101~\citep{feifei2004caltech101} & Object recognition & 6,085 & 102 & mean per class & \cmark & \cmark \\
     & CIFAR-10~\citep{krizhevsky2009cifar} & Visual recognition & 10,000 & 10 & accuracy & \cmark & \cmark \\
     & CIFAR-100~\citep{krizhevsky2009cifar} & Visual recognition & 10,000 & 100 & accuracy & \cmark & \cmark \\
     & CLEVR Counts~\citep{johnson2017clevr,zhai2019vtab} & Counting & 15,000 & 8 & accuracy &  & \cmark \\
     & CLEVR Distance~\citep{johnson2017clevr,zhai2019vtab} & Distance prediction & 15,000 & 6 & accuracy &  & \cmark \\
     & Country211~\citep{radford2021clip,thomee2016yfcc100m} & Geolocation & 21,100 & 211 & accuracy & \cmark & \cmark \\
     & Describable Textures~\citep{cimpoi2014dtd} & Texture classification & 1,880 & 47 & accuracy & \cmark & \cmark \\
     & EuroSAT~\citep{helber2019eurosat,zhai2019vtab} & Satellite imagery recognition & 5,400 & 10 & accuracy & \cmark & \cmark \\
     & FGVC Aircraft~\citep{maji2013fgvc} & Aircraft recognition & 3,333 & 100 & mean per class & \cmark & \cmark \\
     & Food-101~\citep{bossard2014food101} & Food recognition & 25,250 & 101 & accuracy & \cmark & \cmark \\
     & GTSRB~\citep{stallkamp2011gtsrb} & Traffic sign recognition & 12,630 & 43 & accuracy & \cmark & \cmark \\
     & ImageNet 1k~\citep{russakovsky2015imagenet} & Visual recognition & 50,000 & 1,000 & accuracy & \cmark & \cmark \\
     & ImageNet Sketch~\citep{wang2019imagenetsketch} & Visual recognition & 50,889 & 1,000 & accuracy & \cmark &  \\
     & ImageNet v2~\citep{recht2019imagenetv2} & Visual recognition & 10,000 & 1,000 & accuracy & \cmark &  \\
     & ImageNet-A~\citep{hendrycks2021natural} & Visual recognition & 7,500 & 200 & accuracy & \cmark &  \\
     & ImageNet-O~\citep{hendrycks2021natural} & Visual recognition & 2,000 & 200 & accuracy & \cmark &  \\
     & ImageNet-R~\citep{hendrycks2021manyfaces} & Visual recognition & 30,000 & 200 & accuracy & \cmark &  \\
     & KITTI Vehicle Distance~\citep{geiger2012kitti,zhai2019vtab} & Distance prediction & 711 & 4 & accuracy &  & \cmark \\
     & MNIST~\citep{lecun1998mnist} & Digit recognition & 10,000 & 10 & accuracy & \cmark & \cmark \\
     & ObjectNet~\citep{barbu2019objectnet} & Visual recognition & 18,574 & 113 & accuracy & \cmark &  \\
     & Oxford Flowers-102~\citep{nilsback2008flowers102} & Flower recognition & 6,149 & 102 & mean per class & \cmark & \cmark \\
     & Oxford-IIIT Pet~\citep{parkhi2012pets,zhai2019vtab} & Pet classification & 3,669 & 37 & mean per class & \cmark & \cmark \\
     & Pascal VOC 2007~\citep{everingham2007pascalvoc} & Object recognition & 14,976 & 20 & accuracy & \cmark & \cmark \\
     & PatchCamelyon~\citep{veeling2018patchcamelyon,zhai2019vtab} & Metastatic tissue cls. & 32,768 & 2 & accuracy &  & \cmark \\
     & Rendered SST2~\citep{zhai2019vtab} & Sentiment classification & 1,821 & 2 & accuracy & \cmark & \cmark \\
     & RESISC45~\citep{cheng2017resisc45,zhai2019vtab} & Satellite imagery recognition & 6,300 & 45 & accuracy & \cmark & \cmark \\
     & Stanford Cars~\citep{krause2013cars} & Vehicle recognition & 8,041 & 196 & accuracy & \cmark & \cmark \\
     & STL-10~\citep{coates2011stl10} & Visual recognition & 8,000 & 10 & accuracy & \cmark & \cmark \\
     & SUN397~\citep{xiao2010sun} & Scene recognition & 108,754 & 397 & accuracy & \cmark &  \\
     & SVHN~\citep{netzer2011svhn,zhai2019vtab} & Digit recognition & 26,032 & 10 & accuracy & \cmark & \cmark \\
     & FER2013~\citep{goodfellow2013fer2013} & Facial expression recognition & N/A & N/A & accuracy &  & \cmark \\
     & Diabetic Retinopathy~\citep{zhai2019vtab} & Medical image cls. & N/A & N/A & accuracy &  & \cmark \\
     & DMLab~\citep{zhai2019vtab} & 3D scene recognition & N/A & N/A & accuracy &  & \cmark \\
     & dSprites Orientation~\citep{matthey2017dsprites,zhai2019vtab} & Orientation prediction & N/A & N/A & accuracy &  & \cmark \\
     & dSprites X Position~\citep{matthey2017dsprites,zhai2019vtab} & Position prediction & N/A & N/A & accuracy &  & \cmark \\
     & dSprites Y Position~\citep{matthey2017dsprites,zhai2019vtab} & Position prediction & N/A & N/A & accuracy &  & \cmark \\
     & SmallNORB Azimuth~\citep{zhai2019vtab} & Azimuth prediction & N/A & N/A & accuracy &  & \cmark \\
     & SmallNORB Elevation~\citep{zhai2019vtab} & Elevation prediction & N/A & N/A & accuracy &  & \cmark \\
    \midrule
    Retrieval & Flickr30k~\citep{young2014flickr30k} & Image and text retrieval & 1,000 & N/A & R@1 & \cmark &  \\
     & MSCOCO~\citep{lin2014mscoco} & Image and text retrieval & 5,000 & N/A & R@1 & \cmark &  \\
    \bottomrule
  \end{tabular}%
  }%
\end{table}

\begin{table}[htbp]
  \caption{Dataset groups used for per-group analysis. Groups are not exhaustive of all evaluation datasets. \rev{Table~\ref{tab:eval-datasets} lists which datasets are evaluated zero-shot and which with linear probes.}}
  \label{tab:dataset-groups}
  \centering
  \small
  \begin{tabular}{ll}
    \toprule
    Group & Datasets \\
    \midrule
    ImageNet & ImageNet-1k \\
    ImageNet Shifts & ImageNet Sketch, ImageNet v2, ImageNet-A, ImageNet-O, ImageNet-R \\
    Fine-grained & Stanford Cars, Country211, FGVC Aircraft, Food-101, DTD, \\
                 & Flowers-102, Oxford-IIIT Pets, RESISC45, \rev{SUN397}, Diabetic Retinopathy, PatchCamelyon \\
    Retrieval & Flickr30k, MSCOCO \\
    \bottomrule
  \end{tabular}
\end{table}

\FloatBarrier
\section{Additional results}

\subsection{Robustness of the zero-shot degradation to the training seed}
\label{sec:seed_variance}
To confirm that the zero-shot degradation from oversizing the text encoder is not an artifact of a single training run,
we retrain the pair with the smallest degradation in the grid, Base--Tiny (best) versus Base--Giant (degraded),
at three random seeds and test the per-seed paired difference using a paired $t$-test (Table~\ref{tab:seed-variance}).

\begin{table}[htbp]
  \caption{Smallest zero-shot degradation is robust to the training seed.
  Average zero-shot performance 
  for the Base--Tiny (best) and Base--Giant (largest text encoder; degraded) 
  configurations, each trained at three seeds (123, 321, 0).
  $\Delta$ is the per-seed paired difference;
  a one-sided paired $t$-test on $\Delta$ (H$_0$: $\Delta \le 0$)
  gives $t(2){=}3.29$, $p{=}0.04$,
  Cohen's $d_z{=}1.901$, and a one-sided $95\%$ CI of $[+0.09,\ \infty)$. \rev{Seed 0 is the main-grid run.}}
  \label{tab:seed-variance}
  \centering
  \small
  \setlength{\tabcolsep}{6pt}
  \begin{tabular}{lccc}
    \toprule
    Seed & Base--Tiny (\%) & Base--Giant (\%) & $\Delta$ (pp; Tiny $-$ Giant) \\
    \midrule
    $123$    & $38.70$ & $37.68$ & $+1.02$ \\
    $321$    & $38.23$ & $37.15$ & $+1.08$ \\
    $0$ & $38.25$ & $37.93$ & $+0.32$ \\
    \midrule
    Mean & $38.39$ & $37.59$ & $+0.81$ \\
    SD & $0.27$ & $0.40$ & $0.42$  \\
    \bottomrule
  \end{tabular}
\end{table}
\FloatBarrier
\subsection{Zero-shot and linear probe performance for sub-groups}

\begin{figure}[htbp]
  \centering
  \includegraphics[width=\linewidth]{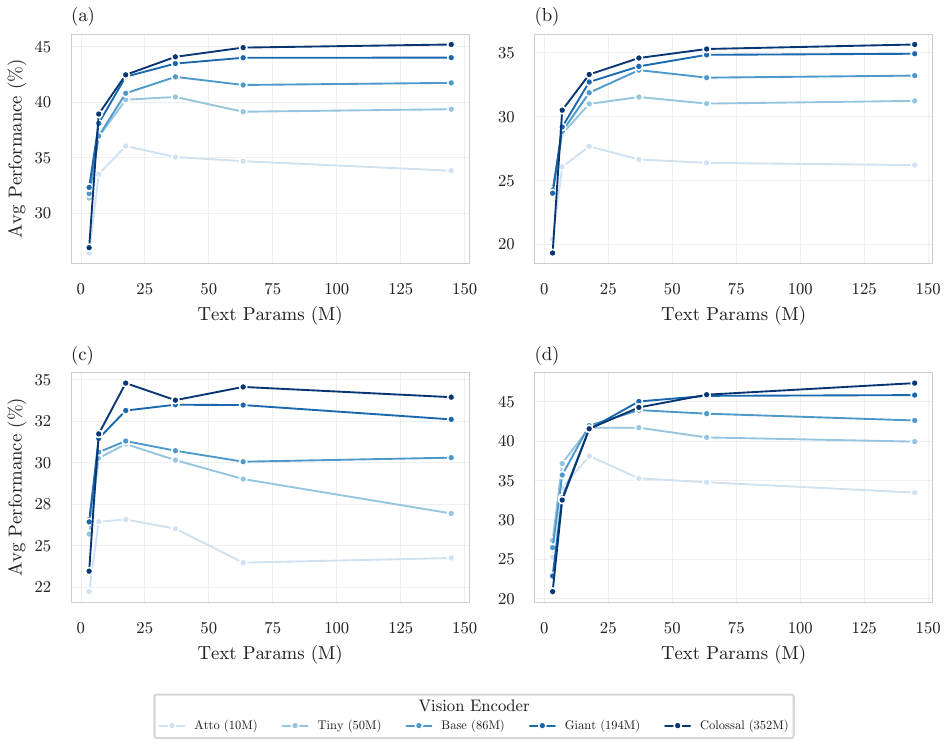}
  \caption{The zero-shot peak-and-degradation pattern holds across groups and is most pronounced for fine-grained tasks.
  Average zero-shot performance (y-axis; higher is better) versus text encoder size (x-axis), colored by vision encoder, for CC12M models split by groups defined in Table~\ref{tab:dataset-groups}:
  (a)~ImageNet, (b)~ImageNet distribution shifts,
  (c)~fine-grained, and
  (d)~retrieval.}
  \label{fig:zs-by-group}
\end{figure}

\begin{figure}[htbp]
  \centering
  \includegraphics[width=\linewidth]{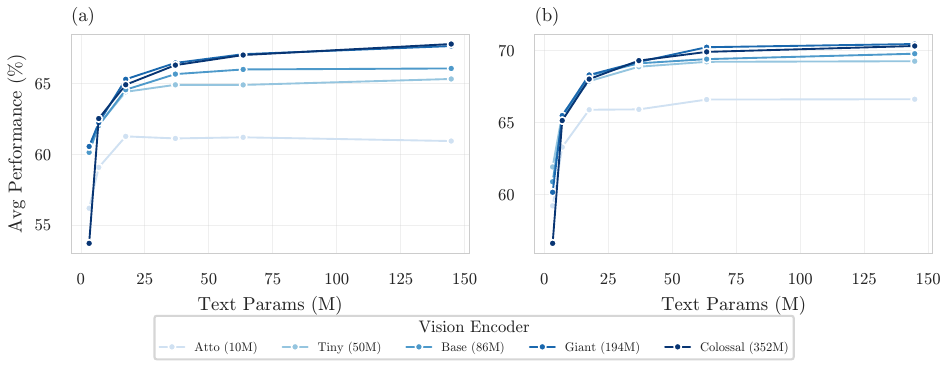}
  \caption{Linear probe scaling shows no degradation across groups.
  Average linear probe performance (y-axis; higher is better) versus text encoder size (x-axis), colored by vision encoder, for CC12M models split by groups defined in Table~\ref{tab:dataset-groups}:
  (a)~ImageNet and
  (b)~fine-grained.}
  \label{fig:lp-by-group}
\end{figure}

The main text reports aggregate scaling trends averaged across all downstream datasets.
To understand how these trends vary across distinct groups, we compute the average zero-shot and linear probe
performance for each dataset group in Table~\ref{tab:dataset-groups}. 
The results are shown in Figures~\ref{fig:zs-by-group} and~\ref{fig:lp-by-group}.

\FloatBarrier
\subsection{CLIP learning curves}

\begin{figure}[htbp]
  \centering
  \includegraphics[width=\linewidth]{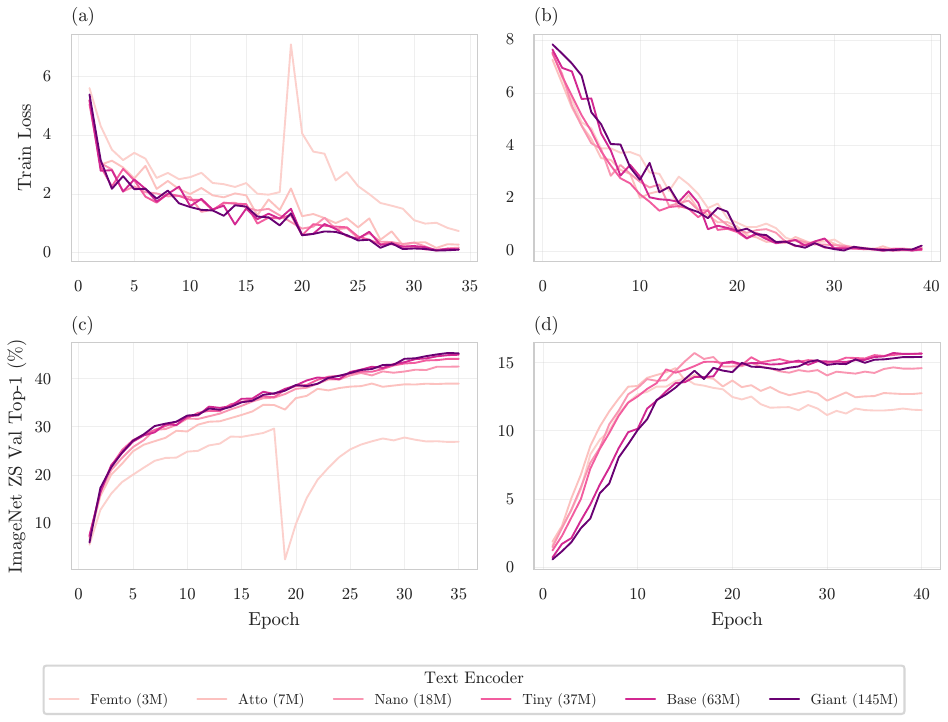}
  \caption{Using the Colossal vision encoder leads to strong overfitting on CC3M but generalizes on CC12M.
  Training loss (top row; a,~b) and ImageNet zero-shot validation top-1 accuracy (bottom row; c,~d; higher is better) versus epoch (x-axis) for the
  Colossal vision encoder pre-trained on (a,~c)~CC12M and (b,~d)~CC3M,
  colored by text encoder.}
  \label{fig:learning_curves_colossal}
\end{figure}

\begin{figure}[htbp]
  \centering
  \includegraphics[width=\linewidth]{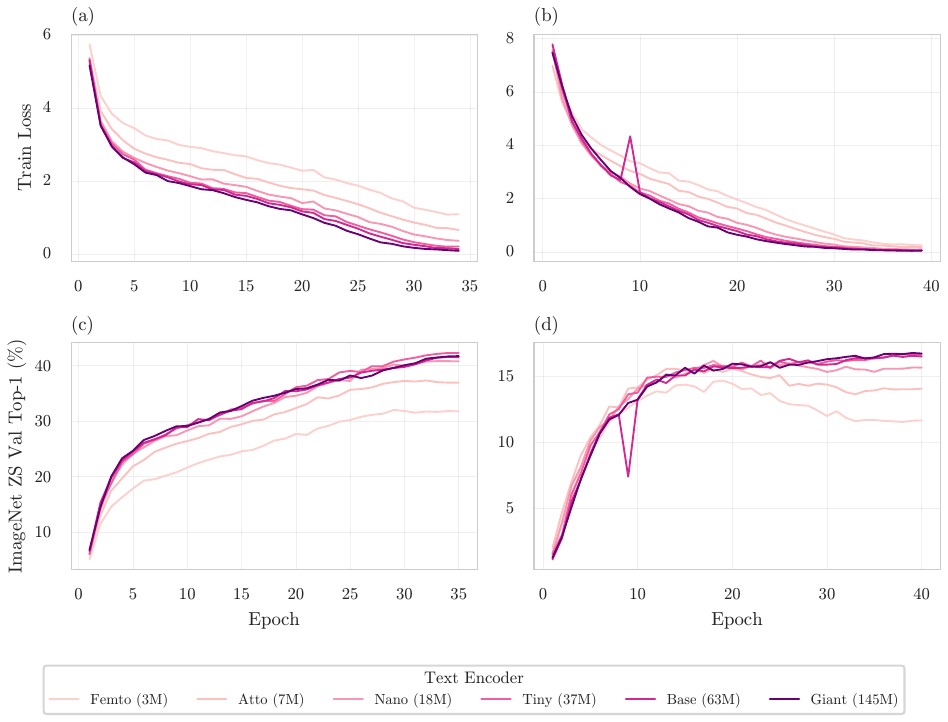}
  \caption{Using the Base vision encoder leads to milder overfitting on CC3M.
  Training loss (top row; a,~b) and ImageNet zero-shot validation top-1 accuracy (bottom row; c,~d; higher is better) versus epoch (x-axis) for the
  Base vision encoder pre-trained on (a,~c)~CC12M and (b,~d)~CC3M,
  colored by text encoder.}
  \label{fig:learning_curves_base}
\end{figure}

To diagnose overfitting to the CC3M and CC12M pre-training datasets, we visualize the training loss and ImageNet zero-shot validation accuracy curves for the Colossal (Fig.~\ref{fig:learning_curves_colossal})
and Base (Fig.~\ref{fig:learning_curves_base}) vision encoders paired with all text encoders. 
On CC3M the Colossal's vision encoder training loss continues to decrease when paired with any text encoder, 
but the validation accuracy saturates or decreases after 20 epochs.
On CC12M, both the training loss and validation accuracy continue to improve throughout training.
The overfitting pattern on CC3M is less pronounced for the Base vision encoder.

\FloatBarrier
\subsection{Text encoder scaling drives linear probe gains without degradation}
\label{sec:linear_probe}

\begin{figure}[htbp]
  \centering
  \includegraphics[width=\linewidth]{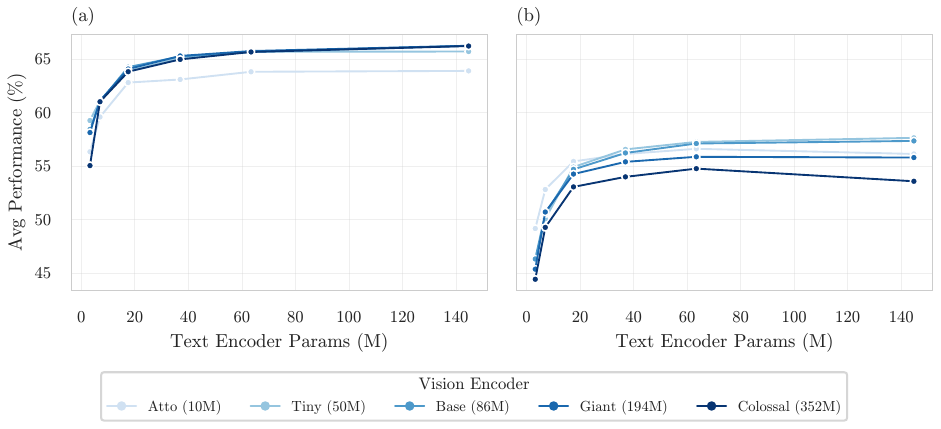}
  \caption{Unlike zero-shot, scaling the text encoder does not degrade linear probe performance.
  Average linear probe performance (y-axis; higher is better), averaged over 31 downstream tasks,
  as a function of text encoder size (x-axis) on (a)~CC12M and (b)~CC3M, colored by vision encoder.}
  \label{fig:lp_scaling}
\end{figure}

Linear probe evaluation provides complementary insights to the zero-shot setting. During inference it relies only on the vision
encoder, but the text encoder is still used throughout training.
We evaluate the linear probe performance of all the models across 31 downstream datasets (Fig.~\ref{fig:lp_scaling}). 
Two patterns differ from the zero-shot setting and hold consistently across both datasets.
First, scaling the text encoder does not degrade performance; in fact, it even improves it when paired with any of the three largest vision encoders.
Second, CC12M models using a vision encoder larger than Atto show that scaling the text encoder leads to larger improvements than scaling the vision encoder.
As in the zero-shot setting, the smallest text encoder (Femto; 3M parameters) leads to poor and unpredictable performance.
These patterns are consistent across downstream task groups
(Fig.~\ref{fig:lp-by-group}): for both ImageNet and fine-grained tasks,
scaling the text encoder yields saturation or monotonic improvement, but no degradation.

On CC3M, the two largest vision encoders show reduced linear probe performance due to overfitting on the smaller dataset, 
consistent with the same effect observed in the zero-shot setting.

\FloatBarrier

\subsection{Predicting zero-shot performance from encoder size}
\label{sec:regression_analysis}
Table~\ref{tab:scaling-model-comparison} reports fit statistics for the scaling models discussed in Section~\ref{sec:allocation}, including the number of free parameters and leave-one-out cross-validated $R^2$.
All models are fit to predict the average zero-shot performance across all tasks using the last epoch of the 30 grid models trained on CC12M dataset.
\begin{table}[htbp]
  \caption{A saturating-ceiling model with a capacity-mismatch penalty best predicts zero-shot performance from encoder size.
  Fit statistics for five models of average zero-shot performance across the 30 CC12M architectures (final-epoch checkpoint), 
  where $N_v$ and $N_t$ are the vision and text encoder parameter counts (millions): number of free parameters, in-sample $R^2$, adjusted $R^2$, and leave-one-out cross-validated $R^2$ (LOOCV; higher is better).
  \rev{The leave-one-vision-out $R^2$ (LOVO) instead holds out all models that share a vision encoder in turn.}
  The mismatch-penalty model (last row) is defined in Eq.~\ref{eq:ceiling}.
  Best metrics are bolded.}
  \label{tab:scaling-model-comparison}
  \centering
  \small
  \setlength{\tabcolsep}{4pt}
  \resizebox{\linewidth}{!}{%
  \begin{tabular}{llccccc}
    \toprule
    Model & Functional form & \# params & $R^2$ & Adj.\ $R^2$ & LOOCV $R^2$ & \rev{LOVO $R^2$} \\
    \midrule
    linear-log: total params & $\beta_0 + \beta_1\log_{10}(N_v{+}N_t)$ & 2 & 0.336 & 0.312 & 0.240 & \rev{0.269} \\
    linear-log: vision + text & $\beta_0 + \beta_1\log_{10}N_v + \beta_2\log_{10}N_t$ & 3 & 0.639 & 0.613 & 0.523 & \rev{0.485} \\
    linear-log: vision + text + interaction & $\beta_0 + \beta_1\log_{10}N_v + \beta_2\log_{10}N_t + \beta_3(\log_{10}N_v)(\log_{10}N_t)$ & 4 & 0.724 & 0.693 & 0.543 & \rev{0.556} \\
    saturating ceiling & $(\beta_0 + \beta_1\log_{10}N_v)(1 - e^{-kN_t})$ & 3 & 0.880 & 0.871 & 0.826 & \rev{0.755} \\
    saturating ceiling with mismatch penalty & $(\beta_0 + \beta_1\log_{10}N_v)(1 - e^{-kN_t})\exp\!\left(-c\left(\log_{10}\tfrac{N_v}{N_t} \rev{{}- r_0}\right)^2\right)$ & \rev{5} & \textbf{\rev{0.971}} & \textbf{\rev{0.967}} & \textbf{\rev{0.946}} & \textbf{\rev{0.912}} \\
    \bottomrule
  \end{tabular}%
  }

\end{table}
\FloatBarrier
\subsection{\texorpdfstring{\rev{Retrieval on a held-out CC12M split}}{Retrieval on a held-out CC12M split}}
\label{sec:overfitting_full}

\begin{figure}[htbp]
  \centering
  \includegraphics[width=\linewidth]{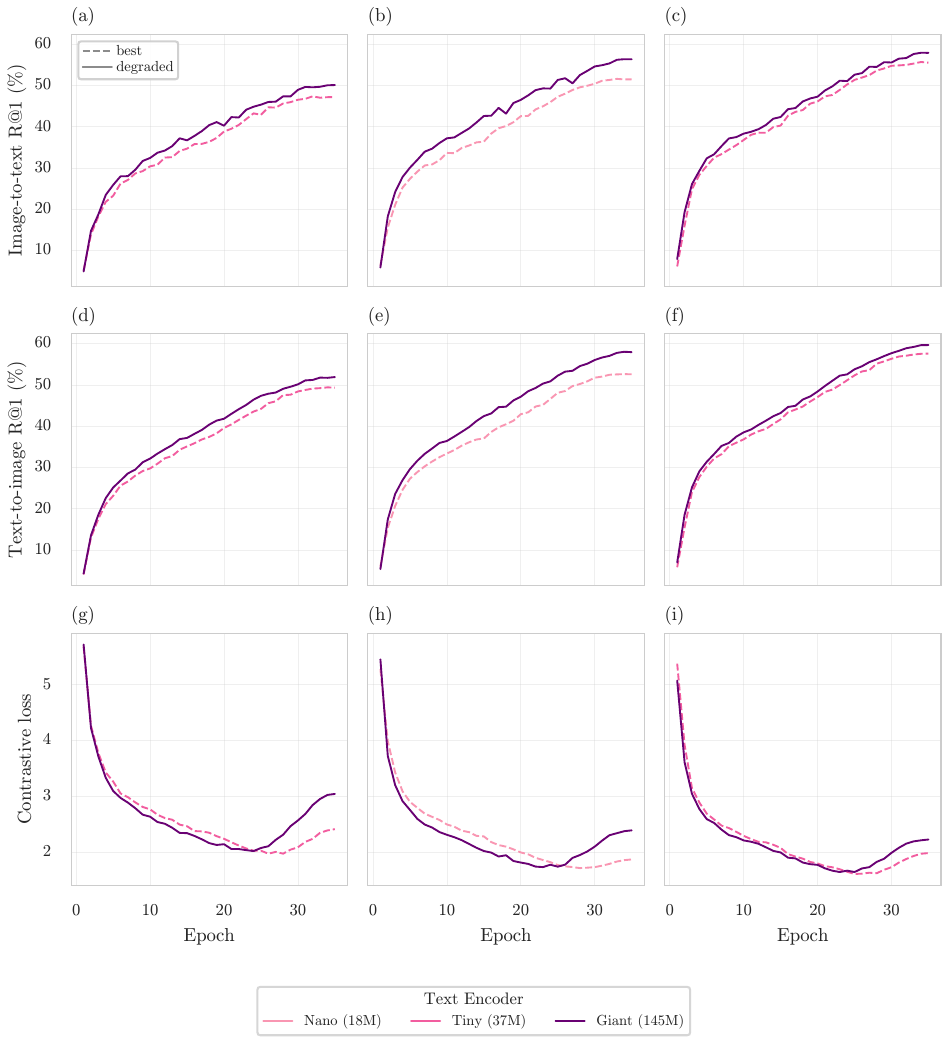}
  \caption{\rev{Larger text encoders achieve higher retrieval recall on held-out CC12M pairs, despite a higher contrastive loss at the end of training.
  Metrics on 24,576 CC12M pairs held out from training, per training epoch (x-axis): (a--c)~image-to-text and (d--f)~text-to-image R@1 (higher is better), and (g--i)~contrastive loss (lower is better).
  (a, d, g)~Atto--Tiny (dashed line; best configuration for that vision encoder) and Atto--Giant (solid line; degraded configuration for that vision encoder).
  (b, e, h)~Tiny--Nano (dashed line; best configuration) and Tiny--Giant (solid line; degraded configuration).
  (c, f, i)~Base--Tiny (dashed line; best configuration) and Base--Giant (solid line; degraded configuration).}}
  \label{fig:heldout_recall}
\end{figure}

\rev{We retrained the Atto, Tiny and Base vision encoders on CC12M with 24,576 pairs (about 0.2\%) held out, each paired with its best text encoder (Tiny, Nano and Tiny, respectively) and with the Giant text encoder (one run each).
At the final epoch, the Giant text encoder has a higher held-out contrastive loss (3.05 vs.\ 2.42 for Atto; 2.40 vs.\ 1.88 for Tiny; 2.23 vs.\ 1.99 for Base; Fig.~\ref{fig:heldout_recall}(g--i)) but also a higher learned logit scale (91.7 vs.\ 82.4; 90.8 vs.\ 80.2; 89.3 vs.\ 84.5).
Retrieval recall (R@1), which does not depend on the temperature, is higher for the Giant text encoder at every epoch after the first (Fig.~\ref{fig:heldout_recall}(a--f)), even though its zero-shot performance is lower (Fig.~\ref{fig:zs_scaling}(a)).}

\FloatBarrier
\subsection{Modality-specific weight decay}
\label{sec:wd_ablation_full}
Table~\ref{tab:wd-ablation-full} reports the average zero-shot performance of training the Base--Giant
and Giant--Giant vision encoders using modality-specific weight decay.
Raising the text encoder's weight decay improves zero-shot performance while raising the vision encoder's degrades it.
\rev{The weight-decay runs in this section, including the runs with the default weight decay ($\lambda_t{=}\lambda_v{=}0.5$) that they are compared against (Tables~\ref{tab:wd-ablation-full} and~\ref{tab:degraded-fixed}), were trained as a separate batch from the main grid. The only differences are the number of GPUs (32 or 64, versus 32 for the main grid) and OpenCLIP's \texttt{-{}-local-loss} and \texttt{-{}-gather-with-grad} options, used for memory efficiency; the global batch size is unchanged (8,192). These options give numerically identical results, so differences from the main-grid results for the same configurations (e.g., Base--Giant, Table~\ref{tab:seed-variance}) reflect run-to-run variation: the number of GPUs changes how the training data are split across GPUs and data-loading workers, and GPU kernels are not deterministic, so even runs with the same seed differ. All comparisons in this section are made within this batch, except the controls with smaller text encoders (Table~\ref{tab:wd-controls}), which use the main-grid setup and are compared with the main-grid models.}

\begin{table}[htbp]
  \caption{Modality-specific weight-decay ablations.
  \rev{(a)}~Average zero-shot performance (higher is better) for every weight-decay ablation run of the Base--Giant and Giant--Giant configurations.
  $\lambda_t$ and $\lambda_v$ are the per-encoder weight decays (baseline $\lambda_t{=}\lambda_v{=}0.5$).
  Within each configuration, the two best-performing weight-decay settings are bolded.
  \rev{(b)~Two-sided sign tests for the comparisons in Section~\ref{sec:mswd}, over tasks and after grouping possibly related tasks (Section~\ref{sec:eval_protocols}); ``Wins'' counts the tasks on which the first configuration is better. The ablations are compared at $\lambda{=}1$.}
  \rev{(c)~The same modality-specific weight-decay intervention for configurations with the Base vision encoder paired with smaller text encoders: average zero-shot performance (higher is better) at the baseline text weight decay ($\lambda_t{=}0.5$) and raised to $\lambda_t{=}1.0$ (vision $\lambda_v{=}0.5$ throughout). ``Improved tasks'' counts datasets ($n$ of $N$) where $\lambda_t{=}1.0$ beats the baseline; $p$ is a two-sided sign test, and ``Grouped'' repeats it after grouping possibly related tasks (Section~\ref{sec:eval_protocols}). $\Delta$ is computed before rounding.}}
  \label{tab:wd-ablation-full}
  \centering
  \small
  \begin{subtable}[t]{\textwidth}
    \centering
    \caption{\rev{Average zero-shot performance.}}
  \begin{tabular}{lrrr}
    \toprule
    Architecture & $\lambda_t$ & $\lambda_v$ & Avg Performance (\%) \\
    \midrule
    Base--Giant & 0.5 & 0.5 & 37.0 \\
     & 0.75 & 0.5 & 38.4 \\
     & 1.0 & 0.5 & \textbf{39.3} \\
     & 0.5 & 0.75 & 36.8 \\
     & 0.5 & 1.0 & 36.3 \\
     & 0.75 & 0.75 & 38.3 \\
     & 1.0 & 1.0 & \textbf{38.5} \\
    \midrule
    Giant--Giant & 0.5 & 0.5 & 39.8 \\
     & 0.75 & 0.5 & 40.4 \\
     & 1.0 & 0.5 & \textbf{41.5} \\
     & 0.5 & 0.75 & 39.7 \\
     & 0.5 & 1.0 & 39.3 \\
     & 0.75 & 0.75 & \textbf{41.6} \\
     & 1.0 & 1.0 & 40.3 \\
    \bottomrule
  \end{tabular}
  \end{subtable}

  \vspace{1.5em}

  \begin{subtable}[t]{\textwidth}
    \centering
    \caption{\rev{Sign tests.}}
    \label{tab:wd-sign-tests}
    \rev{\begin{tabular}{lrrrr}
      \toprule
      Comparison & Wins ($n/N$) & $p$-value & Grouped ($n/N$) & $p$-value \\
      \midrule
      Base--Giant: text-only vs.\ vision-only weight decay & 25/30 & <0.001 & 17/21 & 0.007 \\
      Base--Giant: text-only vs.\ joint weight decay & 23/30 & 0.005 & 16/21 & 0.027 \\
      Giant--Giant: text-only vs.\ vision-only weight decay & 23/30 & 0.005 & 15/21 & 0.078 \\
      Giant--Giant ($\lambda_t{=}1.0$) vs.\ Giant--Tiny & 22/30 & 0.016 & 13/21 & 0.383 \\
      Giant--Giant ($\lambda_t{=}1.0$) vs.\ Colossal--Giant & 17/30 & 0.585 & 12/21 & 0.664 \\
      \bottomrule
    \end{tabular}}
  \end{subtable}

  \vspace{1.5em}

  \begin{subtable}[t]{\textwidth}
    \centering
    \caption{\rev{Controls with smaller text encoders.}}
    \label{tab:wd-controls}
    \rev{\begin{tabular}{lccrrrrr}
      \toprule
      & \multicolumn{2}{c}{Avg Performance (\%)} & & & & & \\
      \cmidrule(lr){2-3}
      Architecture & $\lambda_t{=}0.5$ & $\lambda_t{=}1.0$ & $\Delta$ (pp) & Improved tasks ($n/N$) & $p$-value & Grouped ($n/N$) & $p$-value \\
      \midrule
      Base--Femto & 31.6 & 31.3 & $-$0.3 & 17/30 & 0.585 & 12/21 & 0.664 \\
      Base--Atto & 36.2 & 35.8 & $-$0.4 & 14/30 & 0.856 & 9/21 & 0.664 \\
      Base--Nano & 38.1 & 38.1 & 0.0 & 11/30 & 0.200 & 10/21 & 1.000 \\
      Base--Tiny & 38.2 & 38.9 & +0.7 & 15/30 & 1.000 & 13/21 & 0.383 \\
      Base--Base & 37.7 & 39.5 & +1.7 & 24/30 & 0.001 & 17/21 & 0.007 \\
      \bottomrule
    \end{tabular}}
  \end{subtable}
\end{table}

\begin{figure}[htbp]
  \centering
  \includegraphics[scale=1]{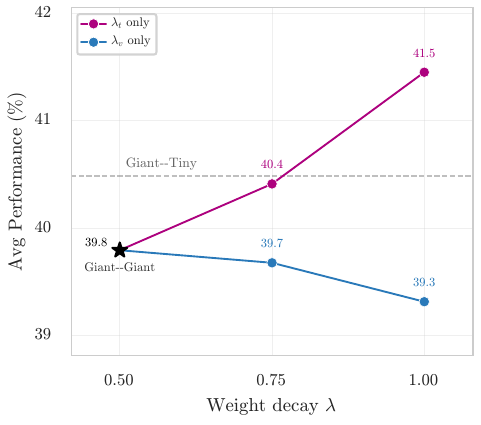}
  \caption{Weight-decay vision-text asymmetry replicates on the Giant--Giant
  configuration. Average zero-shot performance (y-axis; higher is better)
  as weight decay is increased on the text encoder only ($\lambda_t$) or the vision encoder only ($\lambda_v$), both emanating from the Giant--Giant baseline with joint weight decay of 0.5.
  The dashed line marks the best configuration using the Giant vision encoder (Giant--Tiny).}
  \label{fig:wd_ablation_giant}
\end{figure}
  
\begin{figure}[htbp]
  \centering
  \includegraphics[width=\linewidth]{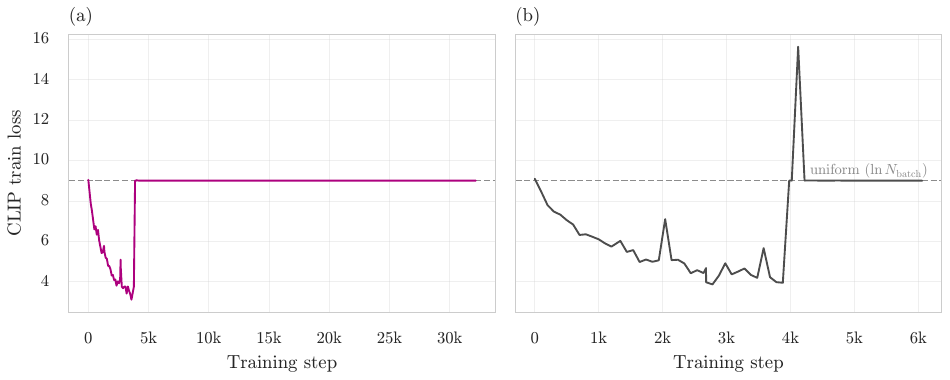}
  \caption{Learning collapses when weight decay is too strong.
  CLIP training loss (y-axis) versus training step (x-axis) for the Base--Giant configuration on
  CC12M with weight decay raised to $\lambda{=}1.5$ on (a)~the text encoder only and (b)~both encoders
  jointly.}
  \label{fig:wd_collapse}
\end{figure}

Raising the weight decay to $\lambda{=}1.5$ destabilizes
training. As shown in Figure~\ref{fig:wd_collapse}, the CLIP training loss decreases normally, then
spikes and locks at $\ln N_{\mathrm{batch}}$, the contrastive loss of a model that outputs a uniform
distribution over the training batch ($N_{\mathrm{batch}}{=}8{,}192$). 

As shown in Section~\ref{sec:mswd}, applying modality-specific weight decay to the degraded configurations improves average zero-shot performance at the expense of training loss \rev{(Table~\ref{tab:degraded-fixed})}.

\begin{table}[htbp]
  \caption{Modality-specific weight decay improves every degraded configuration.
  Average zero-shot performance (higher is better) for each vision encoder paired with the oversized Giant text encoder, at the baseline text weight decay ($\lambda_t{=}0.5$) and raised to $\lambda_t{=}1.0$ (vision $\lambda_v{=}0.5$ throughout).
  ``Improved tasks'' counts datasets ($n$ of $N$) where $\lambda_t{=}1.0$ beats the baseline; $p$ is a two-sided sign test.
  \rev{``Grouped'' repeats the test after grouping possibly related tasks (Section~\ref{sec:eval_protocols}).}}
  \label{tab:degraded-fixed}
  \centering
  \small
  \begin{tabular}{lccrrrrr}
    \toprule
    & \multicolumn{2}{c}{Avg Performance (\%)} & & & & & \\
    \cmidrule(lr){2-3}
    Architecture & $\lambda_t{=}0.5$ & $\lambda_t{=}1.0$ & $\Delta$ (pp) & Improved tasks ($n/N$) & $p$-value & \rev{Grouped ($n/N$)} & \rev{$p$-value} \\
    \midrule
    Atto--Giant & 30.3 & 33.0 & +2.7 & 25/30 & <0.001 & \rev{16/21} & \rev{0.027} \\
    Tiny--Giant & 35.2 & 37.9 & +2.7 & 27/30 & <0.001 & \rev{18/21} & \rev{0.001} \\
    Base--Giant & 37.0 & 39.3 & +2.3 & 24/30 & 0.001 & \rev{17/21} & \rev{0.007} \\
    Giant--Giant & 39.8 & 41.5 & +1.7 & 21/29 & 0.024 & \rev{15/20} & \rev{0.041} \\
    \bottomrule
  \end{tabular}

\end{table}

\FloatBarrier
\subsection{Additional embedding geometry analysis}
\label{sec:add_geometry}

\subsubsection{Uniformity and Alignment}

Given image and text inputs $x_I$ and $x_T$, the image encoder $f$ and text encoder $g$ (including their projection heads) map inputs to $\ell_2$-normalized embeddings on the unit hypersphere.
Let $p_{\text{pos}}$ be the distribution of positive (matched) image--text pairs,
and $p_{\text{image}}$ and $p_{\text{text}}$ be the marginal distributions over images and texts, respectively:

\[
 \mathcal{L}_{\text{align}}(f, g; \alpha) \triangleq \mathbb{E}_{(x_I, x_T) \sim p_{\text{pos}}} \left[ \| f(x_I) - g(x_T) \|_2^{\alpha} \right]
\]

\[
 \mathcal{L}_{\text{uniform}}^{\text{image}}(f; \tau) \triangleq \ln \mathbb{E}_{\substack{x_I, x_I' \overset{\text{i.i.d.}}{\sim} p_{\text{image}}}} \left[ e^{-\tau \| f(x_I) - f(x_I') \|_2^2} \right]
\]

\[
 \mathcal{L}_{\text{uniform}}^{\text{text}}(g; \tau) \triangleq \ln \mathbb{E}_{\substack{x_T, x_T' \overset{\text{i.i.d.}}{\sim} p_{\text{text}}}} \left[ e^{-\tau \| g(x_T) - g(x_T') \|_2^2} \right]
\]

where we set $\alpha = 2$ for alignment and $\tau = 2$ for uniformity, following the original formulation in~\citet{wang2020alignment}.

\begin{figure}[htbp]
  \centering
  \includegraphics[width=\linewidth]{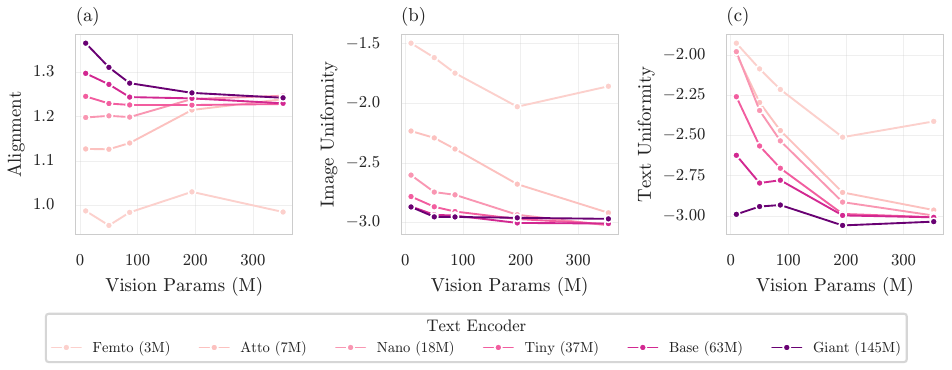}
  \caption{Complementary to Figure~\ref{fig:geometry_text}, scaling the vision encoder improves alignment when paired with larger text encoders but worsens it with the smallest ones.
  (a)~Alignment, (b)~image uniformity, and
  (c)~text uniformity (lower is better for all three; y-axes) of MSCOCO embeddings from the CC12M pre-trained models,
  as a function of vision encoder size (x-axis) and colored by text encoder.}
  \label{fig:geometry_vision}
\end{figure}

Section~\ref{sec:geometry} examines alignment and uniformity as a function of text encoder size.
Here we present the complementary view: the same metrics plotted as a function of vision encoder size (Fig.~\ref{fig:geometry_vision}).
This view helps visualize the impact of scaling the vision encoder on alignment and uniformity.

\subsubsection{Predicting zero-shot performance from uniformity and alignment}

\begin{figure}[htbp]
  \centering
  \includegraphics[scale=1]{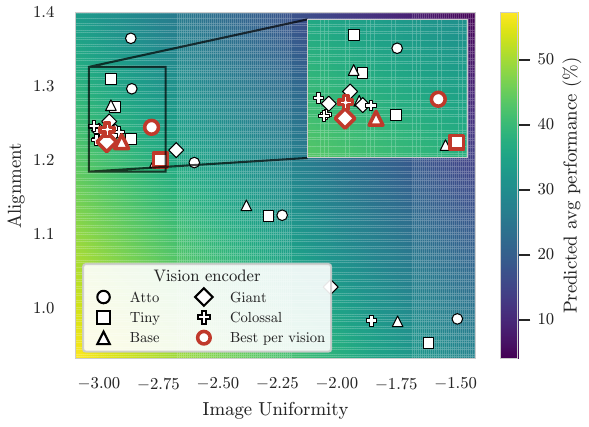}
  \caption{The best models have similar image uniformity and alignment, while degraded models have worse alignment without a compensating gain in image uniformity.
  Predicted average zero-shot performance (color; brighter is higher) across the image uniformity (x-axis; $n{=}200$ grid points) and alignment (y-axis; $n{=}200$ grid points) plane,
  from a linear model fit on the MSCOCO geometry metrics using CC12M models to extract the embeddings.
  Markers show models by vision encoder, with the best text pairing per vision encoder outlined in red. The inset magnifies the boxed region.}
  \label{fig:geometry_landscape}
\end{figure}

\begin{figure}[htbp]
  \centering
  \includegraphics[width=\linewidth]{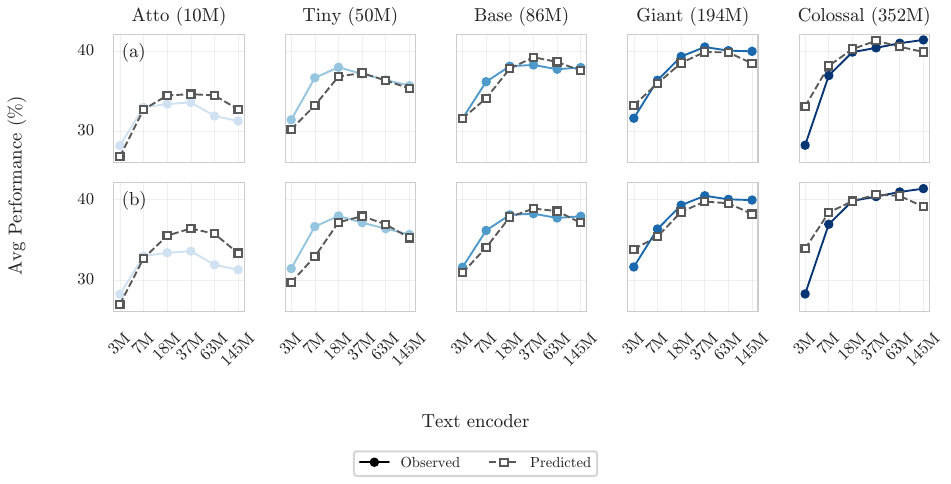}
  \caption{A geometry-only linear model generalizes to unseen vision encoders, capturing the peak-and-degradation pattern.
  For each vision encoder, the model regressing average zero-shot performance on image uniformity and alignment is fit on the four
  other vision encoders and used to predict the held-out one (leave-one-vision-out). Each panel shows the observed (solid line, filled circles,
  coloured by vision encoder) and predicted (dashed line, open grey squares) average zero-shot performance (y-axis; higher is better)
  versus text encoder size (x-axis).
  Each column is a vision encoder while rows give the held-out set used to measure geometry: (a)~Flickr30k and (b)~MSCOCO.}
  \label{fig:geometry_lovo}
\end{figure}

Table~\ref{tab:geometry-prediction}
reports the in-sample and leave-one-out cross-validated $R^2$ of linear models regressing zero-shot performance on
standardized geometry predictors. For this analysis we use all trained CC12M models, including
those used for the modality-specific weight decay ablations, and we perform the cross-validation separately on the MSCOCO and Flickr30k held-out sets.
No single metric predicts zero-shot performance well, but image uniformity and
alignment together explain most of the variance; adding
text uniformity does not help, as it is collinear with image uniformity. To evaluate
whether the regression model generalizes across vision encoders and whether it captures the peak and degradation pattern, 
we perform a leave-one-vision-out (LOVO) analysis, where the linear model is fit using data from all vision encoders except one and used to predict the zero-shot performance on data from the held-out vision encoder.
\rev{Every run built on the held-out vision encoder is excluded from the fit, its modality-specific weight decay variants as well as its grid models, so no architecture appears in both the fitted and the predicted set, and only grid models are predicted.}
Figure~\ref{fig:geometry_lovo} shows that the model generalizes well to unseen vision encoders and captures the overall curve.
For the geometry landscape plot (Fig.~\ref{fig:geometry_landscape}),
we create a grid of 200$\times$200 points across the image uniformity and alignment plane,
and use the linear model to predict the average zero-shot performance at each point.

\begin{table}[htbp]
  \centering
  \caption{Image uniformity and alignment together predict zero-shot performance across models.
  In-sample $R^2$, adjusted $R^2$, and leave-one-out cross-validated $R^2$ (LOOCV; higher is better) for linear models regressing average zero-shot performance on standardized geometry predictors (alignment, image/text uniformity), fit across the CC12M models.
  Geometry is measured on two probe sets ($n{=}38$ models each; MSCOCO, Flickr30k).
  \rev{Brackets give 95\% percentile bootstrap confidence intervals for the LOOCV $R^2$ (10{,}000 resamples over models).}
  (a)~All predictor combinations; the best value in each column is in bold.
  \rev{(b)~The best combination (image uniformity + alignment), recomputed with MSCOCO and Flickr30k datasets removed from the evaluation suite.}}
  \label{tab:geometry-prediction}

  \begin{subtable}[t]{\textwidth}
    \centering
    \caption{Predictive power of each combination of geometry metrics.}
    \label{tab:geometry-prediction-all}
    \small
    \setlength{\tabcolsep}{3pt}
    \begin{tabular}{lccccccc}
      \toprule
      & & \multicolumn{3}{c}{MSCOCO} & \multicolumn{3}{c}{Flickr30k} \\
      \cmidrule(lr){3-5} \cmidrule(lr){6-8} 
      Predictors & \# params & $R^2$ & Adj.\ $R^2$ & LOOCV & $R^2$ & Adj.\ $R^2$ & LOOCV \\
      \midrule
      Image uniformity & 2 & 0.490 & 0.476 & 0.445 \rev{{\scriptsize[0.10, 0.67]}} & 0.524 & 0.511 & 0.482 \rev{{\scriptsize[0.16, 0.69]}} \\
      Alignment & 2 & 0.150 & 0.127 & 0.022 \rev{{\scriptsize[-0.41, 0.29]}} & 0.098 & 0.072 & -0.047 \rev{{\scriptsize[-0.42, 0.18]}} \\
      Text uniformity & 2 & 0.347 & 0.328 & 0.275 \rev{{\scriptsize[-0.11, 0.52]}} & 0.435 & 0.420 & 0.381 \rev{{\scriptsize[0.01, 0.63]}} \\
      Image uniformity + alignment & 3 & 0.847 & \textbf{0.838} & \textbf{0.810} \rev{{\scriptsize[0.68, 0.90]}} & 0.901 & \textbf{0.895} & \textbf{0.877} \rev{{\scriptsize[0.78, 0.95]}} \\
      All three & 4 & \textbf{0.848} & 0.835 & 0.794 \rev{{\scriptsize[0.65, 0.90]}} & \textbf{0.901} & 0.893 & 0.867 \rev{{\scriptsize[0.76, 0.94]}} \\
      \bottomrule
    \end{tabular}
  \end{subtable}

  \vspace{1.5em}

  \begin{subtable}[t]{\textwidth}
    \centering
    \caption{\rev{Image uniformity $+$ alignment, excluding MSCOCO and Flickr30k from the evaluation suite.}}
    \label{tab:geometry-prediction-noleak}
    \small
    \setlength{\tabcolsep}{3pt}
    \begin{tabular}{lccccccc}
      \toprule
      & & \multicolumn{3}{c}{\rev{MSCOCO}} & \multicolumn{3}{c}{\rev{Flickr30k}} \\
      \cmidrule(lr){3-5} \cmidrule(lr){6-8} 
      \rev{Zero-shot evaluation suite} & \rev{\# tasks} & \rev{$R^2$} & \rev{Adj.\ $R^2$} & \rev{LOOCV} & \rev{$R^2$} & \rev{Adj.\ $R^2$} & \rev{LOOCV} \\
      \midrule
      \rev{All tasks} & \rev{30} & \rev{0.847} & \rev{0.838} & \rev{0.810 {\scriptsize[0.68, 0.90]}} & \rev{0.901} & \rev{0.895} & \rev{0.877 {\scriptsize[0.78, 0.95]}} \\
      \rev{Excluding MSCOCO and Flickr30k} & \rev{26} & \rev{0.839} & \rev{0.829} & \rev{0.805 {\scriptsize[0.66, 0.91]}} & \rev{0.897} & \rev{0.891} & \rev{0.875 {\scriptsize[0.77, 0.95]}} \\
      \midrule
      \rev{$\Delta$} & \rev{$-4$} & \rev{$-0.008$} & \rev{$-0.009$} & \rev{$-0.005$} & \rev{$-0.004$} & \rev{$-0.005$} & \rev{$-0.002$} \\
      \bottomrule
    \end{tabular}
  \end{subtable}

\end{table}

\subsubsection{Impact of modality-specific weight decay on embedding geometry}

\begin{figure}[htbp]
  \centering
  \includegraphics[width=\linewidth]{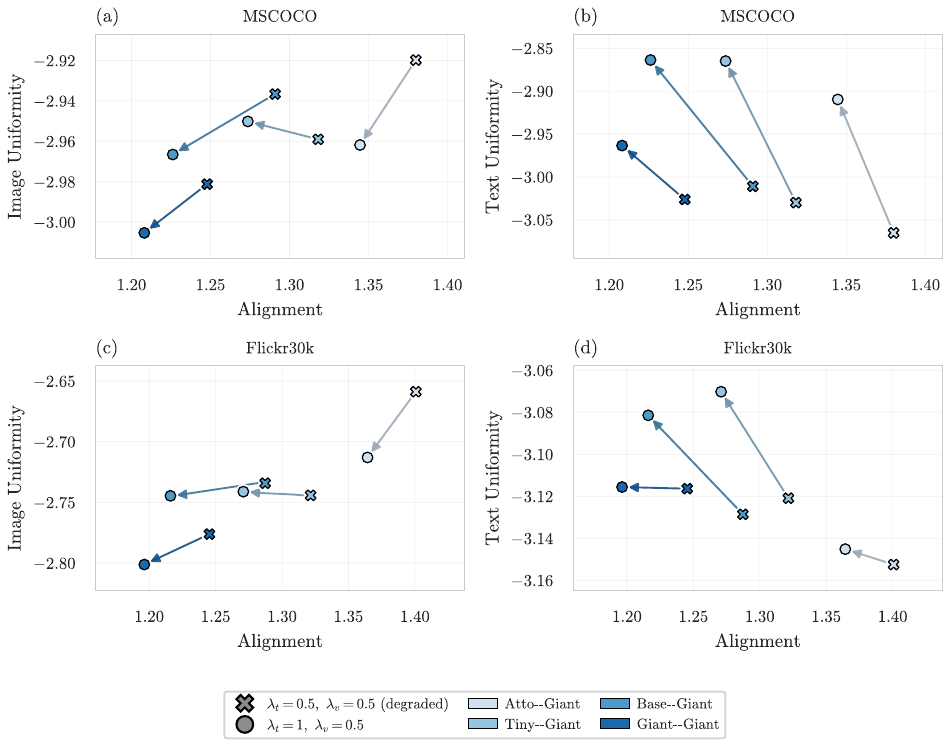}
  \caption{Increasing text weight decay improves alignment and image uniformity.
  Alignment (x-axis) versus uniformity (y-axis; lower is better for both) for each vision
  encoder paired with the Giant text encoder, going from the degraded configuration
  ($\lambda_t{=}0.5$, $\lambda_v{=}0.5$; crosses) to the configuration with modality-specific weight decay ($\lambda_t{=}1$, $\lambda_v{=}0.5$;
  circles). Rows: (a,b)~MSCOCO and (c,d)~Flickr30k held-out sets. Columns: (a,c)~image uniformity
   and (b,d)~text uniformity (y-axis).}
  \label{fig:degraded_fixed_geometry_grid}
\end{figure}

Figure~\ref{fig:degraded_fixed_geometry_grid} shows the change in
alignment and uniformity when going from each degraded configuration to its counterpart using modality-specific weight decay, for both MSCOCO and Flickr30k held-out
sets and for both image and text uniformity. Panel~(a) corresponds to Figure~\ref{fig:degraded_fixed_geometry}.

\subsubsection{RankMe}
\begin{figure}[htbp]
  \centering
  \includegraphics[width=\linewidth]{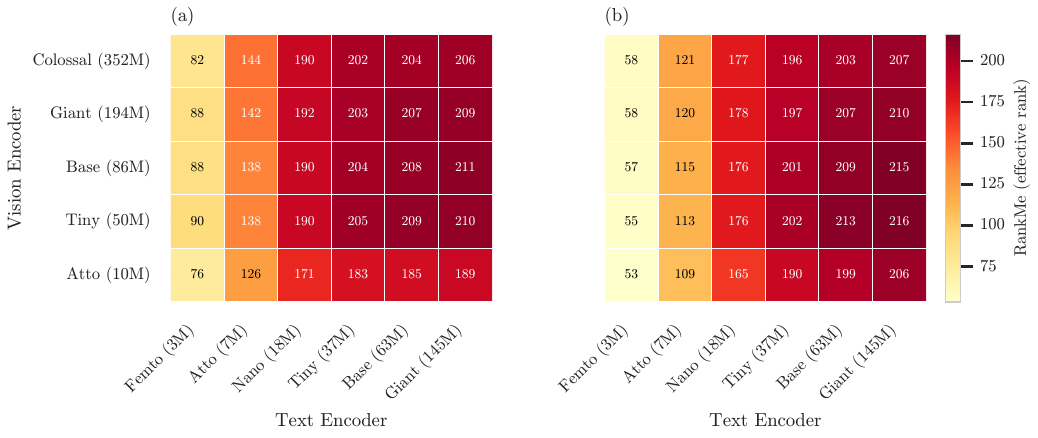}
  \caption{Scaling the text encoder raises the effective rank (RankMe) of both image and text embeddings, whereas the vision encoder size has little effect.
  (a)~Image and (b)~text RankMe for the 30 architectures on CC12M training data, crossing five vision encoders (y-axis) with six text encoders (x-axis).}
  \label{fig:rankme_heatmap}
\end{figure}

We also compute RankMe~\citep{garrido2023rankme} for the image and text embeddings (Fig.~\ref{fig:rankme_heatmap}), 
which provides a label-free estimate of the effective dimensionality of a representation matrix. RankMe is computed on a random sample of 25,600 images from the CC12M training data, 
matching the sample size and methodology used in the original work~\citep{garrido2023rankme}.

Given an embedding matrix $\boldsymbol{Z}$, RankMe is defined as:
\[
  \text{RankMe}(\boldsymbol{Z}) = \exp\!\left(-\sum_{k=1}^{\min(N,K)} p_k \ln p_k\right), \quad \text{with} \quad p_k = \frac{\sigma_k(\boldsymbol{Z})}{\|\sigma(\boldsymbol{Z})\|_1} + \epsilon,
\]
where $\sigma_k(\boldsymbol{Z})$ denotes the $k$-th singular value of $\boldsymbol{Z}$.

Scaling the text encoder increases the effective rank of both image and
text embeddings substantially, 
while RankMe is significantly less
sensitive to the size of the vision encoder (Fig.~\ref{fig:rankme_heatmap}).
This pattern closely mirrors the linear probe scaling behavior
(Fig.~\ref{fig:lp_scaling}), where the text encoder is the
dominant driver of performance. This is also consistent with ~\citet{garrido2023rankme}, which shows that RankMe correlates with linear
probe downstream performance.

\subsubsection{Modality gap}
\begin{figure}[htbp]
  \centering
  \includegraphics[scale=1]{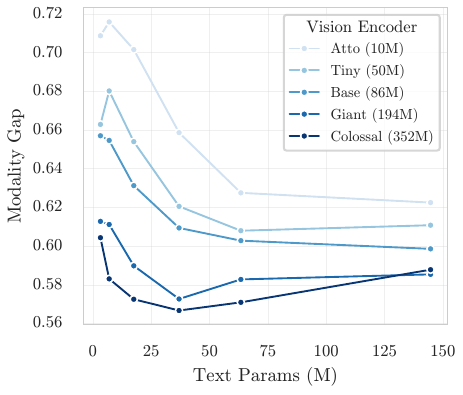}
  \caption{\rev{As the text encoder grows, the modality gap mostly decreases, with a small uptick at the largest text encoders for the two largest vision encoders.}
  Modality gap (y-axis) of MSCOCO embeddings as a function of text encoder size (x-axis) for CC12M models, colored by vision encoder.}
  \label{fig:modality_gap}
\end{figure}

The modality gap is a phenomenon observed in contrastive vision-language models where image and text embeddings cluster in distinct cones of the embedding space~\citep{liang2022mind}.
It is measured as the distance between the centroids of the image and text embedding distributions:
\[
  \Delta_{\text{gap}} = \left\| \frac{1}{n}\sum_{i=1}^{n} f(x_I^{(i)}) - \frac{1}{n}\sum_{i=1}^{n} g(x_T^{(i)}) \right\|_2
\]
The modality gap decreases roughly monotonically as either encoder is scaled up,
though the two largest vision encoders exhibit a small uptick at the largest text encoder sizes (Fig.~\ref{fig:modality_gap}).
Overall, the modality gap does not exhibit the non-monotonic pattern observed in zero-shot performance,
indicating that it may not be sufficient to explain zero-shot downstream performance.

\subsection{\texorpdfstring{\rev{Dropout in the text encoder}}{Dropout in the text encoder}}
\label{sec:dropout}
\rev{To test whether the recovery is specific to weight decay, we retrained the four degraded configurations (Atto--Giant, Tiny--Giant, Base--Giant and Giant--Giant) with dropout ($p{=}0.1$) in the text encoder at the baseline weight decay ($\lambda_t{=}\lambda_v{=}0.5$).
We apply dropout to the token embeddings after adding the positional embeddings, to the outputs of the attention and MLP blocks, and also apply attention dropout.
Dropout improves all four configurations on most tasks, by 3.1, 3.3, 2.1 and 2.3 pp (Table~\ref{tab:dropout}), similar to raising the text encoder's weight decay (Table~\ref{tab:degraded-fixed}).
This result strengthens the view that the degradation comes from oversized text encoders, and that different regularization strengths for different modalities improve out-of-distribution generalization.}

\begin{table}[htbp]
  \caption{\rev{Dropout in the text encoder also improves the degraded configurations.
  Average zero-shot performance of the four degraded configurations retrained with dropout ($p{=}0.1$) in the text encoder, all using the baseline weight decay.
  ``Improved tasks'' counts tasks ($n$ of $N$) where the dropout model is better; $p$ is a two-sided sign test, and ``Grouped'' repeats it after grouping possibly related tasks (Section~\ref{sec:eval_protocols}). $\Delta$ is computed before rounding.}}
  \label{tab:dropout}
  \centering
  \small
  \rev{\begin{tabular}{lccrrrrr}
    \toprule
    & \multicolumn{2}{c}{Avg Performance (\%)} & & & & & \\
    \cmidrule(lr){2-3}
    Architecture & No dropout & Dropout & $\Delta$ (pp) & Improved tasks ($n/N$) & $p$-value & Grouped ($n/N$) & $p$-value \\
    \midrule
    Atto--Giant & 31.3 & 34.4 & +3.1 & 27/30 & <0.001 & 18/21 & 0.001 \\
    Tiny--Giant & 35.7 & 38.9 & +3.3 & 25/30 & <0.001 & 17/21 & 0.007 \\
    Base--Giant & 37.9 & 40.1 & +2.1 & 24/30 & 0.001 & 16/21 & 0.027 \\
    Giant--Giant & 39.9 & 42.2 & +2.3 & 27/30 & <0.001 & 19/21 & <0.001 \\
    \bottomrule
  \end{tabular}}
\end{table}

\subsection{\texorpdfstring{\rev{Model efficiency}}{Model efficiency}}
\label{sec:efficiency}
\rev{Table~\ref{tab:efficiency} reports the parameters, FLOPs, latency, throughput and memory of every configuration in the main grid.
We report forward FLOPs per sample (one $224\times224$ image and one 77-token caption), counted with PyTorch's FLOP counter (\texttt{torch.utils.flop\_counter}), as in OpenCLIP's model profile~\citep{cherti2023openclip}. The count covers all matrix multiplications and convolutions, including the attention products $QK^\top$ and $\mathrm{softmax}(\cdot)V$, and counts one multiply--add as two FLOPs; element-wise operations (normalization, softmax, activations) are excluded.}

\rev{We measure latency, throughput and memory on one NVIDIA A100 80GB PCIe GPU with PyTorch~2.12 (eager mode) and fp16 mixed precision, as in training, using random inputs already on the GPU, so data loading is excluded. Latency is the time of one forward pass of the image or text encoder at batch size 1, and inference throughput is 256 divided by the time of one forward pass at batch size 256; each time is the median of 100 calls after 20 warm-up calls, with GPU synchronization before and after each call. Training step time is the median time of a full training step (forward pass, contrastive loss, backward pass and AdamW update) on 256 image--caption pairs, over 20 steps after 5 warm-up steps. Peak memory is the maximum GPU memory occupied by tensors during that step (\texttt{torch.cuda.max\_memory\_allocated}), which includes weights, optimizer state, gradients and activations, but not the CUDA context or memory cached by the allocator. Activation memory is the memory allocated after the forward pass and loss minus the memory allocated before it, i.e., the tensors kept for the backward pass. Memory is reported in GiB.}

\begin{table}[htbp]
  \caption{\rev{Model efficiency of every configuration in the main grid.
  Parameters in millions (``Text non-emb.'' excludes the token and positional embeddings), forward GFLOPs per image and per 77-token caption, latency at batch size 1, inference throughput at batch size 256 in thousands of samples per second, and training step time and memory for a batch of 256 image--caption pairs (Appendix~\ref{sec:efficiency}).}}
  \label{tab:efficiency}
  \centering
  \scriptsize
  \setlength{\tabcolsep}{3pt}
  \rev{\begin{tabular}{lrrrrrrrrrrrr}
    \toprule
    & \multicolumn{3}{c}{Parameters (M)} & \multicolumn{2}{c}{GFLOPs} & \multicolumn{2}{c}{Latency (ms)} & \multicolumn{2}{c}{Throughput (k/s)} & Train step (ms) & \multicolumn{2}{c}{Memory (GiB)} \\
    \cmidrule(lr){2-4} \cmidrule(lr){5-6} \cmidrule(lr){7-8} \cmidrule(lr){9-10} \cmidrule(lr){11-11} \cmidrule(lr){12-13}
    Architecture & Vision & Text & Text non-emb. & Image & Text & Image & Text & Image & Text & Full model & Activation & Peak \\
    \midrule
    Atto--Femto & 9.9 & 3.2 & 0.1 & 4.3 & 0.01 & 6.5 & 0.8 & 10.7 & 288.1 & 80 & 5.5 & 5.9 \\
    Atto--Atto & 9.9 & 7.0 & 0.7 & 4.3 & 0.10 & 6.7 & 2.0 & 10.7 & 124.5 & 83 & 5.7 & 6.1 \\
    Atto--Nano & 9.9 & 17.5 & 4.9 & 4.3 & 0.76 & 6.2 & 3.3 & 10.6 & 50.7 & 95 & 6.6 & 7.1 \\
    Atto--Tiny & 9.9 & 36.9 & 17.9 & 4.3 & 2.82 & 6.6 & 6.0 & 10.6 & 20.3 & 120 & 8.2 & 9.0 \\
    Atto--Base & 9.9 & 63.4 & 38.1 & 4.3 & 5.96 & 6.5 & 6.7 & 10.5 & 12.2 & 148 & 9.9 & 11.0 \\
    Atto--Giant & 9.9 & 144.7 & 106.7 & 4.3 & 16.62 & 6.6 & 8.1 & 10.5 & 5.7 & 221 & 13.8 & 15.9 \\
    \midrule
    Tiny--Femto & 50.2 & 3.2 & 0.1 & 20.6 & 0.01 & 5.3 & 0.8 & 4.2 & 285.3 & 182 & 11.4 & 12.3 \\
    Tiny--Atto & 50.2 & 7.0 & 0.7 & 20.6 & 0.10 & 5.4 & 2.0 & 4.2 & 126.7 & 188 & 11.6 & 12.5 \\
    Tiny--Nano & 50.2 & 17.5 & 4.9 & 20.6 & 0.76 & 5.4 & 3.5 & 4.2 & 49.5 & 201 & 12.4 & 13.4 \\
    Tiny--Tiny & 50.2 & 36.9 & 17.9 & 20.6 & 2.82 & 5.4 & 5.9 & 4.2 & 19.7 & 223 & 14.1 & 15.4 \\
    Tiny--Base & 50.2 & 63.4 & 38.1 & 20.6 & 5.96 & 5.4 & 6.7 & 4.2 & 11.8 & 255 & 15.8 & 17.4 \\
    Tiny--Giant & 50.2 & 144.7 & 106.7 & 20.6 & 16.62 & 5.4 & 8.4 & 4.1 & 5.6 & 329 & 19.7 & 22.3 \\
    \midrule
    Base--Femto & 86.2 & 3.2 & 0.1 & 35.1 & 0.01 & 6.4 & 0.8 & 2.7 & 283.7 & 271 & 16.3 & 17.7 \\
    Base--Atto & 86.2 & 7.0 & 0.7 & 35.1 & 0.10 & 6.4 & 2.0 & 2.7 & 133.5 & 277 & 16.5 & 17.8 \\
    Base--Nano & 86.2 & 17.5 & 4.9 & 35.1 & 0.76 & 6.5 & 3.6 & 2.7 & 48.8 & 288 & 17.4 & 18.8 \\
    Base--Tiny & 86.2 & 36.9 & 17.9 & 35.1 & 2.82 & 6.3 & 5.7 & 2.7 & 19.5 & 316 & 19.0 & 20.7 \\
    Base--Base & 86.2 & 63.4 & 38.1 & 35.1 & 5.96 & 6.5 & 6.9 & 2.7 & 11.7 & 344 & 20.7 & 22.7 \\
    Base--Giant & 86.2 & 144.7 & 106.7 & 35.1 & 16.62 & 6.6 & 8.4 & 2.7 & 5.6 & 414 & 24.6 & 27.6 \\
    \midrule
    Giant--Femto & 194.2 & 3.2 & 0.1 & 79.0 & 0.01 & 10.9 & 0.9 & 1.3 & 275.8 & 552 & 31.3 & 33.9 \\
    Giant--Atto & 194.2 & 7.0 & 0.7 & 79.0 & 0.10 & 11.0 & 2.0 & 1.3 & 118.8 & 561 & 31.5 & 34.0 \\
    Giant--Nano & 194.2 & 17.5 & 4.9 & 79.0 & 0.76 & 10.9 & 3.7 & 1.3 & 48.7 & 582 & 32.3 & 35.0 \\
    Giant--Tiny & 194.2 & 36.9 & 17.9 & 79.0 & 2.82 & 10.9 & 6.0 & 1.3 & 19.5 & 606 & 34.0 & 36.9 \\
    Giant--Base & 194.2 & 63.4 & 38.1 & 79.0 & 5.96 & 10.9 & 6.9 & 1.3 & 11.8 & 624 & 35.7 & 38.9 \\
    Giant--Giant & 194.2 & 144.7 & 106.7 & 79.0 & 16.62 & 10.4 & 8.4 & 1.3 & 5.6 & 703 & 39.6 & 43.8 \\
    \midrule
    Colossal--Femto & 352.4 & 3.2 & 0.1 & 142.3 & 0.01 & 10.9 & 0.8 & 0.7 & 297.7 & 925 & 44.3 & 48.8 \\
    Colossal--Atto & 352.4 & 7.0 & 0.7 & 142.3 & 0.10 & 11.6 & 2.0 & 0.8 & 117.8 & 925 & 44.6 & 48.8 \\
    Colossal--Nano & 352.4 & 17.5 & 4.9 & 142.3 & 0.76 & 11.9 & 3.6 & 0.7 & 48.8 & 949 & 45.4 & 49.8 \\
    Colossal--Tiny & 352.4 & 36.9 & 17.9 & 142.3 & 2.82 & 11.9 & 5.8 & 0.7 & 19.5 & 974 & 47.0 & 51.7 \\
    Colossal--Base & 352.4 & 63.4 & 38.1 & 142.3 & 5.96 & 11.7 & 6.9 & 0.7 & 11.7 & 1013 & 48.7 & 53.7 \\
    Colossal--Giant & 352.4 & 144.7 & 106.7 & 142.3 & 16.62 & 11.5 & 8.5 & 0.7 & 5.5 & 1091 & 52.6 & 58.6 \\
    \bottomrule
  \end{tabular}}
\end{table}

\subsection{\texorpdfstring{\rev{Learned logit scales}}{Learned logit scales}}
\label{sec:logit_scales}
\rev{Table~\ref{tab:logit-scale-grid} reports the learned logit scale (the inverse softmax temperature of the contrastive loss) at the end of training for every model in the main grid, and Table~\ref{tab:logit-scale-reg} for the weight-decay and dropout runs.}

\begin{table}[htbp]
  \caption{\rev{Learned logit scale of every model in the main grid. Final value of the learned logit scale for each vision encoder (rows) and text encoder (columns), trained on (a)~CC12M and (b)~CC3M.}}
  \label{tab:logit-scale-grid}
  \centering
  \small
  \rev{%
  \begin{subtable}[t]{0.49\textwidth}
    \centering
    \caption{CC12M.}
    \setlength{\tabcolsep}{3pt}
    \begin{tabular}{lrrrrrr}
      \toprule
      & \multicolumn{6}{c}{Text encoder} \\
      \cmidrule(lr){2-7}
      Vision encoder & Femto & Atto & Nano & Tiny & Base & Giant \\
      \midrule
      Atto & 71.1 & 68.8 & 74.8 & 82.4 & 87.4 & 92.0 \\
      Tiny & 75.0 & 74.3 & 78.2 & 84.9 & 87.8 & 90.6 \\
      Base & 75.2 & 75.8 & 80.6 & 84.3 & 87.0 & 89.7 \\
      Giant & 80.4 & 80.3 & 82.0 & 83.6 & 85.0 & 86.6 \\
      Colossal & 74.1 & 80.4 & 83.4 & 84.7 & 85.5 & 86.0 \\
      \bottomrule
    \end{tabular}
  \end{subtable}\hfill
  \begin{subtable}[t]{0.49\textwidth}
    \centering
    \caption{CC3M.}
    \setlength{\tabcolsep}{3pt}
    \begin{tabular}{lrrrrrr}
      \toprule
      & \multicolumn{6}{c}{Text encoder} \\
      \cmidrule(lr){2-7}
      Vision encoder & Femto & Atto & Nano & Tiny & Base & Giant \\
      \midrule
      Atto & 57.0 & 60.9 & 68.5 & 80.9 & 85.2 & 85.7 \\
      Tiny & 66.3 & 69.3 & 76.7 & 81.0 & 84.3 & 81.8 \\
      Base & 70.7 & 77.2 & 79.5 & 81.5 & 81.4 & 80.5 \\
      Giant & 75.9 & 79.8 & 80.5 & 80.6 & 79.7 & 77.2 \\
      Colossal & 74.6 & 78.3 & 79.4 & 78.7 & 77.1 & 75.0 \\
      \bottomrule
    \end{tabular}
  \end{subtable}}
\end{table}

\begin{table}[htbp]
  \caption{\rev{Learned logit scale with modality-specific regularization on CC12M. Final learned logit scale at the default weight decay ($\lambda_t{=}\lambda_v{=}0.5$), with the text weight decay raised to $\lambda_t{=}1.0$, and with dropout ($p{=}0.1$) in the text encoder with default weight decay ($\lambda_t{=}\lambda_v{=}0.5$). Dashes indicate no run.}}
  \label{tab:logit-scale-reg}
  \centering
  \small
  \rev{\begin{tabular}{lccc}
    \toprule
    Architecture & $\lambda_t{=}\lambda_v{=}0.5$ & $\lambda_v{=}0.5$, $\lambda_t{=}1.0$ & $\lambda_t{=}\lambda_v{=}0.5$, $p{=}0.1$ \\
    \midrule
    Atto--Giant & 89.7 & 82.2 & 80.3 \\
    Tiny--Giant & 89.2 & 82.5 & 81.9 \\
    Base--Giant & 87.9 & 81.7 & 81.4 \\
    Giant--Giant & 86.9 & 80.4 & 80.5 \\
    \midrule
    Base--Femto & 75.2 & 72.6 & -- \\
    Base--Atto & 75.8 & 73.3 & -- \\
    Base--Nano & 80.6 & 75.7 & -- \\
    Base--Tiny & 84.3 & 77.8 & -- \\
    Base--Base & 87.0 & 79.3 & -- \\
    \bottomrule
  \end{tabular}}
\end{table}

\FloatBarrier
\section{\texorpdfstring{\rev{Extended discussion}}{Extended discussion}}
\label{sec:extended_discussion}

\subsection{\texorpdfstring{\rev{An information-bottleneck view of text-encoder overfitting}}{An information-bottleneck view of text-encoder overfitting}}
\label{sec:ib_discussion}
\rev{The information bottleneck (IB) principle~\citep{tishby2000information} states that a good representation $Z$ of an input $X$ is as informative as possible about a target $Y$ while being as compact as possible:
it minimizes $I(X;Z) - \beta\, I(Z;Y)$, discarding information about the input that is irrelevant to the target.
\citet{ji2025cibr} interpret CLIP's contrastive objective through this lens.
The loss maximizes the information shared by the image and text embeddings, $I(Z_v;Z_t)$, whereas modality-specific information, i.e., what each embedding retains about its own input beyond what the paired input explains ($I(Z_v;X_v \mid X_t)$ and $I(Z_t;X_t \mid X_v)$),
is discarded only implicitly, through the model's finite representational capacity and the regularization inherent in optimization, such as the batch size, the embedding dimension, and gradient noise.
Building on this view, they add an explicit penalty on these conditional terms to the CLIP loss and observe consistent improvements over standard CLIP.}

\rev{Our findings are consistent with this view.
If modality-specific information is discarded only because capacity is limited, enlarging the text encoder relaxes this implicit constraint,
and the text embeddings can retain caption-specific information, such as idiosyncratic phrasing or noise in web captions, that does not help, and may even hurt, cross-modal alignment.
The text-encoder overfitting we observe (Section~\ref{sec:mswd}) would then be a symptom of this retained information.
The embedding geometry points in the same direction: as the text encoder grows, the text embeddings spread more uniformly and their effective rank (RankMe) increases,
while image--text alignment worsens (Section~\ref{sec:geometry}; Fig.~\ref{fig:rankme_heatmap}).}

\rev{Modality-specific weight decay may counteract this effect by regularizing the text encoder, bringing training closer to the IB objective.
Consistent with this, raising $\lambda_t$ in the degraded configurations improves image--text alignment, typically at the expense of text uniformity, i.e., the text embeddings become less spread out (Section~\ref{sec:geometry}; Fig.~\ref{fig:degraded_fixed_geometry_grid}).}

\rev{This is an interpretation rather than a tested explanation or formal proof.}

\end{document}